\documentclass[oneside,a4paper]{article}

\usepackage{bbding,ifsym,pifont}
\usepackage{booktabs,hhline,enumerate}
\usepackage{graphicx,color,subfigure}

\usepackage{bm,subfloat,url}

\usepackage{times}
\usepackage{epsfig}
\usepackage{graphicx}
\usepackage{amsmath,amsthm}
\usepackage{amssymb}
\usepackage{enumerate}
\usepackage[algoruled,vlined]{algorithm2e}
\usepackage{mathtools}
\usepackage{color}
\usepackage{tabu}

\usepackage{multirow}
\newtheorem{assumption}{Assumption}

\usepackage{enumerate,subfigure}

\newtheorem{principle}{Principle}
\newtheorem{remark}{Remark}
\newtheorem{theorem}{Theorem}
\newtheorem{definition}{Definition}
\newtheorem{corollary}{Corollary}
\newtheorem{proposition}{Proposition}

\newtheorem{innercustomthm}{Principle}
\newenvironment{customthm}[1]
  {\renewcommand\theinnercustomthm{#1}\innercustomthm}
  {\endinnercustomthm}

\newcommand{\dt}{\operatorname{d}\!}
\newcommand{\diag}{\operatorname{diag}}

\begin{document}

\begin{center}
	{\Large Towards a Mathematical Understanding of the Difficulty in 
	Learning with Feedforward Neural Networks
	\footnote{Formerly: A Smooth Optimisation Perspective on Designing and
	Training Feedforward Multilayer Perceptrons}
	} \\[6mm]
	{\large Hao~Shen} \\[0.5mm]
	E-mail: hao.shen@fortiss.org \\[2mm]
	fortiss -- Landesforschungsinstitut des Freistaats Bayern, Germany \\[7mm]
\end{center}

\begin{abstract}
	Training deep neural networks for solving machine learning problems is one great 
	challenge in the field, mainly due to its associated 
	optimisation problem being highly non-convex.
	Recent developments have suggested that many training algorithms do not suffer 
	from undesired local minima under certain scenario, and consequently led to 
	great efforts in pursuing mathematical explanations for such observations.
	%
	%
	This work provides an alternative mathematical understanding of the challenge
	from a smooth optimisation perspective.
	By assuming exact learning of finite samples, sufficient conditions 
	are identified via a critical point analysis to ensure any local minimum to 
	be globally mini\-mal as well.
	Furthermore, a state of the art algorithm, known as the Generalised Gauss-Newton (GGN)
	algorithm, is rigorously revisited as an approximate Newton's algorithm, which
	shares the property of being locally quadratically convergent to a global minimum
	under the condition of exact learning.
\end{abstract}

\begin{center}
	\textbf{\small{Index Terms}} \\[2mm]
	Feedforward Multilayer Perceptrons (MLPs), 
	smooth optimisation, 
	critical point analysis, 
	Hessian matrix,
    approximate Newton's method.
\end{center}

\section{Introduction}\label{sec:01}
Deep Neural Networks (DNNs) have been successfully applied to solve 
challenging problems in pattern recognition, computer vision, and speech recognition 
\cite{bish:book96,lecu:nature15,yudo:book15}.
Despite this success, training DNNs is still one of the greatest 
challenges in the field \cite{glor:aistats10}.
In this work, we focus on training the classic feedforward Multi-Layer Perceptrons (MLPs).
It is known that performance of MLPs is highly dependent on various  
factors in a very complicated way.
For example, studies in \cite{horn:nn91,suns:aaai16} identify the topology of MLPs 
as a determinative factor.
Works in \cite{mhas:nips93,glor:aistats10} demonstrate the impact of different 
activation functions to performance of MLPs.
Moreover, a choice of error/loss functions is also shown to be influential as in
\cite{fala:ijcnn99}. 

Even with a well designed MLP architecture, training a specific MLP both effectively and 
efficiently can be as challenging as constructing the network.
The most popular method used in training MLPs is the well-known \emph{backpro\-pagation}
(BP) algorithm \cite{widr:pieee90}. 
Although the classic BP algorithm shares a great convenience of being very 
simple, they can suffer from two major problems, namely, (i) 
existence of undesired local minima, even if global optimality is assumed;
and (ii) slow convergence speed. 
Early works as \cite{sutt:ccss86,roja:book96} argue that such problems with BP 
algorithms are essentially due to their nature of being gradient descent algorithms,
while an associated optimisation problem for MLP training is often highly non-convex
and of large scale.

One major approach to address the problem of undesired local minima in MLP training
is via an error/loss surface analy\-sis \cite{hame:nn98,spri:amai99,chor:aistats15}.
Even for simple tasks, such as the classic XOR problem, the error surface 
analysis is surprisingly complicated and its results can be hard to conclude
\cite{lisb:network91,spri:amai99,spri:tnn99}. 
%
%
Early efforts in \cite{gori:pami92,yuxi:tnn92,yuxi:tnn95} try to 
identify general conditions on the topology of MLPs to eliminate undesired local minima,
i.e., suboptimal local minima.
Unfortunately, these attempts fail to provide complete solutions to general
problems.
On the other hand, although BP algorithms are often thought to be sensitive to 
initialisations \cite{kole:cs90}, recent results reported in \cite{good:iclr15} 
suggest that modern MLP training algorithms can overcome the problem of 
suboptimal local minima conveniently.
Such observations have triggered several very recent efforts to characterise 
global optimality of DNN training \cite{kawa:nips16,nguy:icml17,haef:cvpr17}.

The work in \cite{kawa:nips16} shows that both \emph{deep linear 
networks} and \emph{deep nonlinear networks} with only the Rectified Linear Unit 
(ReLU) function in the hidden layers are free of suboptimal local minima.
The attempted technique is not applicable for analysing \emph{deep 
nonlinear networks} with other activation functions, e.g. the \emph{Sigmoid} 
and the \emph{SoftSign}. 
Recent work in \cite{nguy:icml17} studies a special case of the problem, where the 
training samples are assumed to be linearly independent, i.e., the number of samples 
is smaller than the dimension of the sample space.
Unfortunately, such a strong assumption does not even hold for the classic XOR problem.
Moreover, conditions constructed in its main result (Theorem 3.8 in
\cite{nguy:icml17}), i.e., the existence of non-degenerate global minima, can 
hardly be satisfied in practice (see Remark~\ref{rmk:degenerate} in Section~\ref{sec:04} 
of this paper for details).
%
%
%
Most recently, by restricting the network output and its regularisation to 
be positively homogeneous with respect to the network parameters, the work in 
\cite{haef:cvpr17} develops sufficient conditions to guarantee all local minima to
be globally minimal.
So far, such results only apply to networks with either one hidden layer or multiple 
deep subnetworks connected in parallel.
Moreover, it still fails to explain performance of MLPs with non-positively 
homogeneous activation functions.

To deal with slow convergence speed of the classic BP algorithm,
various modifications have been developed, such as momentum based BP algorithm 
\cite{vogl:bc88}, conjugate gra\-dient algorithm \cite{char:ieepg92}, BFGS algorithm \cite{lequ:icml11}, and ADAptive Moment estimation (ADAM) algorithm \cite{king:iclr15}.
Specifically, a Generalised Gauss-Newton (GGN) algori\-thm proposed 
in \cite{schr:nc02} has demonstrated its prominent performance over many state of 
the art training algorithms in practice \cite{mart:icml10,viny:aistats12}.
Unfortunately, justification for such performance is still mathematically vague 
\cite{mart:arxiv14}.
Another popular solution to deal with slow convergence is to employ Newton's method 
for MLP training. However, an implementation of exact Newton's method is often 
computationally prohibitive.
Hence, approximations of the Hessian matrix are needed to address 
the issue, such as a diagonal approximation structure \cite{batt:nc92}, 
and a block diagonal approximation structure \cite{wang:nn98}.
However, without a complete evaluation of the true Hessian, performance of these 
heuristic approximations is hardly convincing.
Although existing attempts \cite{bish:nc92,mizu:nn08} have characterised the Hessian 
by applying partial derivatives, these results still fail to provide 
further structural information of the Hessian, due to the limitations of 
partial derivatives.

In this work, we provide an alternative mathematical understanding 
of the difficulty in training MLPs from a smooth optimisation perspective.
Sufficient conditions are identified via a critical point analysis to 
ensure all local minima to be globally minimal.
Convergence properties of the GGN algorithm are rigorously 
analysed as an approximate Newton's algorithm.

\section{Learning with MLPs}
\label{sec:02}
Many machine learning tasks can be formulated as a problem of learning a task-specific 
\emph{ground truth mapping} $g^{*} \colon \bm{\mathcal{X}} \to \bm{\mathcal{Y}}$,
where $\bm{\mathcal{X}}$ and $\bm{\mathcal{Y}}$ denote an \emph{input space} and 
an \emph{output space}, respectively.
The problem of interest is to approximate $g^{*}$, given only a finite number of
samples in either $\bm{\mathcal{X}}$ or $\bm{\mathcal{X}} \!\times\! \bm{\mathcal{Y}}$.
When only $T$ samples in $\bm{\mathcal{X}}$ are given, say $\{x_{i}\}_{i=1}^{T} 
\!\subset\! \bm{\mathcal{X}}$, the problem of approximating $g^{*}$
is known as \emph{unsupervised learning}. 
When both input samples and their desired outputs $y_{i} := g^{*}(x_{i})$ are provided, 
i.e., given training samples $\{(x_{i},y_{i})\}_{i=1}^{T} \!\subset\! \bm{\mathcal{X}} 
\!\times\! \bm{\mathcal{Y}}$, the corresponding problem is referred to as 
\emph{supervised learning}. 
In this work, we only focus on the problem of supervised learning.

A popular approach to evaluate learning outcomes is via minimising an empirical 
error/loss function as
\begin{equation}
\label{eq:empirical}
	\widetilde{g} := \operatorname*{argmin}_{g \in \bm{\mathcal{G}}}~\frac{1}{T} 
	\sum\limits_{i=1}^{T} E\big(g(x_{i}), p(x_{i})\big), 
\end{equation}
where $\bm{\mathcal{G}}$ denotes a hypothetical function space, where a minimiser 
$\widetilde{g}$ is assumed to be reachable, and $E \colon \bm{\mathcal{Y}}
\times \bm{\mathcal{Z}} \to \mathbb{R}$ denotes a suitable error function that 
evaluates the estimate $g(x_{i})$ against some task-dependent prior knowledge 
$p(x_{i}) \in \bm{\mathcal{Z}}$.
For supervised learning problems, such prior knowledge is simply the corresponding 
desired output $y_{i}$, i.e., $p(x_{i}) := g^{*}(x_{i})$.
In general, given only a finite number of samples, the ground truth mapping 
$g^{*}$ is hardly possible to be exactly learned as the solution $\widetilde{g}$.
Nevertheless, we can still define a specific scenario of exact learning with
respect to a finite number of samples. 
\begin{definition}[Exact learning]
	Let $g^{*} \colon \bm{\mathcal{X}} \to \bm{\mathcal{Y}}$ be the ground truth
	mapping.
	Given samples $\{x_{i},y_{i}\}_{i=1}^{T} \subset \bm{\mathcal{X}} \times
	\bm{\mathcal{Y}}$, 
	a mapping $\widehat{g} \colon \bm{\mathcal{X}} \!\to\! \bm{\mathcal{Y}}$, 
    which satisfies $\widehat{g}(x_{i}) = g^{*}(x_{i})$
    for all $i = 1, \ldots,T$, 
    %
	%
	is called a \emph{finite exact approximator} of $g^{*}$ based on
	the $T$ samples. 
\end{definition}

For describing MLPs, we denote by $L$ the number of 
layers in an MLP structure, and by $n_{l}$ the number of processing units in 
the $l$-th layer with $l = 1, \ldots, L$.
Specifically, by $l = 0$, we refer it to the input layer.
Let $\sigma \colon \mathbb{R} \to \mathbb{R}$ be a unit activation function, 
which was traditionally chosen to be non-constant, bounded, continuous, 
and monotonically increasing.
Recent choices, e.g. ReLU, SoftPlus, and Bent identity, are unbounded functions.
In this work, we restrict activation functions to be smooth, and denote by 
$\sigma' \colon \mathbb{R} \to \mathbb{R}$ and 
$\sigma'' \colon \mathbb{R} \to \mathbb{R}$ the first and second derivative
of the activation function $\sigma$, respectively. 

For the $(l,k)$-th unit in an MLP architecture, referring to the $k$-th unit in the 
$l$-th layer, we define the corresponding \emph{unit mapping} as
%
%
\begin{equation}
\label{eq:unit_map}
	f_{l,k}(\phi_{l-1}, w_{l,k}) :=
	\sigma\!\left(w_{l,k}^{\top} \phi_{l-1} - b_{l,k}\right),
\end{equation}
where $\phi_{l-1} \in \mathbb{R}^{n_{l-1}}$ denotes the output 
from the $(l-1)$-th layer, $w_{l,k} \in \mathbb{R}^{n_{l-1}}$ and 
$b_{l,k} \in \mathbb{R}$ are respectively a weight vector and a 
bias associated with the $(l,k)$-th unit.
Note, that the bias $b_{l,k}$ is a free variable in general.
However, through our analysis in this work, we fix it as a constant scalar 
as in Eq.~\eqref{eq:unit_map} for the sake of convenience for presentation.
Then, we can simply define the \emph{$l$-th layer mapping} by stacking 
all \emph{unit mappings} in the layer as 
%
%
\begin{equation}
	\!\!\!\!\!F_{l} (W_{l},\! \phi_{l-1}) \!:=\!\! \big[f_{l,1}(w_{l,1},\! \phi_{l-1}), 
	\ldots,\! f_{l,n_{l}}(w_{l,n_{l}},\! \phi_{l-1})]^{\top}\!\!\!,\!\!\! 
\end{equation}
with $W_{l} := [w_{l,1}, \ldots, w_{l,n_{l}}] \in 
\mathbb{R}^{n_{l-1} \times n_{l}}$ being the $l$-th weight matrix.
Specifically, let us denote by $\phi_{0} \in \mathbb{R}^{n_{0}}$ the input, then 
the output at the $l$-th layer is defined as $\phi_{l} := F_{l}(W_{l},\phi_{l-1})$ 
recursively.
Note, that the last layer of an MLP commonly employs the identify map as the activation
function, i.e., $\phi_{L} := W_{L}^{\top} \phi_{L-1}$. 
Finally, by denoting the set of all parameter matrices in the MLP by 
$\bm{\mathcal{W}} := \mathbb{R}^{n_{0} \times n_{1}} \times \ldots \times
\mathbb{R}^{n_{L-1} \times n_{L}}$, we compose all layer-wise mappings to 
define the overall \emph{MLP network mapping} 
\begin{equation}
	\begin{split}
		F \colon \bm{\mathcal{W}} \times \mathbb{R}^{n_{0}} \to &~ \mathbb{R}^{n_{L}}, \\
		(\mathbf{W},\phi_{0}) \mapsto &~ F_{L}(W_{L},\cdot) \circ
		\ldots \circ F_{1}(W_{1},\phi_{0}). 
	\end{split}
\end{equation}
With such a construction, we can define the set of parameterised mappings specified 
by a given MLP architecture as
\begin{equation}
\label{eq:mlp}
	\bm{\mathcal{F}} := \big\{ F(\mathbf{W},\cdot) \colon 
	\mathbb{R}^{n_{0}} \to \mathbb{R}^{n_{L}} 
	\big| \mathbf{W} \in \bm{\mathcal{W}} \big\}. 
\end{equation}
More specifically, we denote by $\bm{\mathcal{F}}(n_{0},\ldots,n_{L})$ the 
MLP architecture specifying the number of units in each layer.

Now, let $\bm{\mathcal{X}} \subseteq \mathbb{R}^{n_{0}}$ and 
$\bm{\mathcal{Y}} \subseteq \mathbb{R}^{n_{L}}$, one can utilise 
an MLP $F(\mathbf{W},\cdot) \in \bm{\mathcal{F}}$ to approximate the 
ground truth mapping $g^{*}$.
Then, an empirical total loss function of MLP based learning can be formulated as 
\begin{equation}
\label{eq:total_loss}
	\mathcal{J}(\mathbf{W}) := \frac{1}{T}  \sum\limits_{i=1}^{T} 
	E\big( F(\mathbf{W},x_{i}), y_{i} \big). 
\end{equation}
If the error function $E( \cdot, \cdot )$ is differentiable in $F(\mathbf{W},x_{i})$,
then the function $\mathcal{J}$ is differentiable in the weights $\mathbf{W}$. 
For the convenience of presentation, in the rest of the paper we denote the
sample-wise loss function by 
\begin{equation}
\label{eq:sample_loss}
	J(\mathbf{W}, x_{i}) := E\big( F(\mathbf{W},x_{i}), y_{i} \big).
\end{equation}
It is important to notice that, even when the error function 
$E( \cdot, \cdot )$ is constructed to be convex with respect to the
first argument, the \emph{total loss function} $\mathcal{J}$ as defined in 
Eq.~\eqref{eq:total_loss} is still \emph{non-convex} in $\mathbf{W}$.
Since the set of parameters $\bm{\mathcal{W}}$ is \emph{unbounded}, when squashing 
activation functions are employed, exploding the norm of any weight matrix will not 
drive the function value of $\mathcal{J}$ to infinity.
Namely, the total loss function $\mathcal{J}$ can be \emph{non-coercive} 
\cite{guel:book10}.
Therefore, the existence and attainability of global minima of $\mathcal{J}$ are 
not guaranteed in general.
However, in the finite sample setting, when appropriate nonlinear activation functions,
such as \emph{Sigmoid} and $\operatorname{tanh}$, are employed in the hidden 
layer, a three-layer MLP with a sufficiently large number of units in 
the hidden layer can achieve exact learning of finite samples
\cite{itoy:acm96,shah:tnn99}.

In the rest of the paper, we assume the existence of an MLP structure that
is capable of realising a finite exact approximator.
\begin{assumption}
\label{ass:exist}
	Let $g^{*} \colon \bm{\mathcal{X}} \to \bm{\mathcal{Y}}$ be the ground truth
	mapping.
	Given $T$ unique samples $\{x_{i}\}_{i=1}^{T} \!\subset\! \bm{\mathcal{X}}$, 
	there exists an MLP architecture $\bm{\mathcal{F}}$, as defined in Eq.~\eqref{eq:mlp},
	and a weight $\mathbf{W}^{*} \in \bm{\mathcal{W}}$, so that the corresponding MLP 
	$F(\mathbf{W}^{*}, \cdot)$ is a finite exact approximator of $g^{*}$.
\end{assumption}

As exact learning of finite samples is assumed, a suitable error function $E$
is critical to ensure its attainability and uniqueness via an optimisation 
procedure.
Optimistically, a finite exact approximator corresponds with a global mini\-mum of
the corresponding total loss function without undesired local minima.
We propose the following assumption as a practical principle of choosing error
function.

\begin{principle}[Choice of error function]
\label{prin:error}
	The error function $E( \cdot, \cdot )$ is differentiable with respect to 
	its first argument.
	If the gradient of $E$ with respect to the first argument vanishes at 
	$\phi_{L} \in \mathbb{R}^{n_{L}}$, i.e., $\nabla_{\!E}(\phi_{L}) = 0$, 
	then $\phi_{L}$ is a global minimum of $E$.
\end{principle}
\begin{remark}
	Typical examples of error function include the classic squared loss, 
	smooth approximations of $\ell_{p}$ norm with $0 < p < 2$, Blake-Zisserman loss, 
	and Cauchy loss \cite{hart:book04}.
	Moreover, by Principle~\ref{prin:error}, the weights $\mathbf{W}^{*}$ as assumed 
	in Assum\-ption~\ref{ass:exist} is a global minimiser of the total loss function 
	$\mathcal{J}$. 
\end{remark}

\section{Critical point analysis of MLP training}
\label{sec:03}
%

In order to develop a gradient descent algorithm to mini\-mise the cost function 
$\mathcal{J}$ as in Eq.~\eqref{eq:total_loss}, the derivative 
of all \emph{unit mappings} are building blocks for our computation.
We define the derivative of the activation function $\sigma$ in the $(l,k)$-th unit as 
\begin{equation}
	f'_{l,k}(w_{l,k}, \phi_{l-1}) :=
	\sigma'\left(w_{l,k}^{\top} \phi_{l-1} - b_{l,k}\right),
\end{equation}
and the collection of all derivatives of activation functions in the $l$-th layer 
as 
\begin{equation}
	F'_{l}(W_{l}, \!\phi_{l-1}) \!:=\!\! 
	\big[f'_{_{\!l,1}}\!(w_{_{l,1}}\!, \phi_{_{l-1}}\!), \ldots,
	f'_{_{\!l,n_{l}}}\!(w_{_{l,n_{l}}}\!, \phi_{_{l-1}}\!)\big]^{\top}\!\!\!.\!
\end{equation}
%
For simplicity, we denote by $\phi'_{l} := F'_{l}(W_{l}, \phi_{l-1}) 
\in \mathbb{R}^{n_{l}}$.
%
%
We apply the chain rule of multivariable derivative to compute the directional 
derivative of $J$ with respect to $W_{l}$ in direction 
$H_{l} \in \mathbb{R}^{n_{l-1} \times n_{l}}$ as 
\begin{align*}
	\!\operatorname{D}\!J(W_{l}) \!\cdot\! H_{l} \!= & \operatorname{D}\!E(\phi_{L}) \cdot
	\operatorname{D}_{2}\!F_{L}(W_{L},\phi_{L-1}) \cdot \ldots \cdot \tag{11} \\
	\cdot & \operatorname{D}_{2}\!F_{l+1}(W_{l+1},\!\phi_{l}) \cdot
	\operatorname{D}_{1}\!F_{l}(W_{l},\!\phi_{l-1}) \cdot H_{l},\!\!
	\setcounter{equation}{11}
\end{align*}
where $\operatorname{D}_{1}\!F_{l}(W_{l},\phi_{l-1}) \!\cdot\! H_{l}$ and 
$\operatorname{D}_{2}\!F_{l}(W_{l},\phi_{l-1}) \!\cdot\! h_{l-1}$ refer to the directional derivative 
of $F_{l}$ with respect to the first and the second argument,
respectively.
Explicitly, the first derivative of $F_{l}$, i.e.,
$\operatorname{D}_{1}\!F_{l}(W_{l},\phi_{l-1}) \colon$ $\mathbb{R}^{n_{l-1}\times n_{l}} \to 
\mathbb{R}^{n_{l}}$ as a linear map acting on direction 
$H_{l} \in \mathbb{R}^{n_{l-1}\times n_{l}}$, is 
\begin{equation}
	\operatorname{D}_{1}\!F_{l}(W_{l},\phi_{l-1}) \!\cdot\! H_{l} = 
	\operatorname{diag}(\phi'_{l}) H_{l}^{\top} \phi_{l-1},
\end{equation}
where the operator $\diag(\cdot)$ puts a vector into a diagonal matrix form, and
the first derivative of $F_{l}$ with respect to the second parameter
in direction $h_{l-1} \in \mathbb{R}^{n_{l-1}}$ as
\begin{equation}
	\operatorname{D}_{2}\!F_{l}(W_{l},\phi_{l-1}) \!\cdot\! h_{l-1} = 
	\operatorname{diag}(\phi'_{l}) W_{l}^{\top} h_{l-1}.
\end{equation}
\begin{algorithm}[t!]
\caption{BP algorithm for supervised learning (batch learning).} 
\label{algo:01} 
\SetAlgoNoLine
	\SetKwHangingKw{IN}{Input~:}
	\SetKwHangingKw{Sa}{Step~1:}
	\SetKwHangingKw{Sb}{Step~2:}
	\SetKwHangingKw{Sc}{Step~3:}
	\SetKwHangingKw{Sd}{Step~4:}
	\SetKwHangingKw{Se}{Step~5:}
	\SetKwHangingKw{Sf}{Step~6:}
	\SetKwHangingKw{Sg}{Step~7:}
	\SetKwHangingKw{OUT}{\hspace{-1.6mm}Output:}

	\IN{Samples $(x_{i},y_{i})_{i=1}^{T}$, an MLP architecture 
		$F_{n_{1},\ldots,n_{L}}$, and initial weights $\mathbf{W}$ }
		
	\OUT{Accumulation point $\mathbf{W}^{*}$ \vspace{0mm}}
				
	\Sa{For $i = 1,\ldots,T$, feed samples $x_{i}$ through the 
		MLP to compute \vspace{1mm}
		
		\hspace{10.5mm} (1) $\phi_{l}^{(i)}\!$ and ${\phi'_{l}}^{(i)}\!$ for 
			$l = 1, \ldots, L$; \vspace{1mm}
		
		\hspace{10.5mm} (2) $\omega_{L}^{(i)} := 
			\operatorname{diag}({\phi'_{L}}^{(i)}) \nabla_{E}(\phi_{L}^{(i)}) \in 
			\mathbb{R}^{n_{L}}$ \vspace{1mm}
		}
	
	\Sb{For $l = L,\ldots,1$, compute \vspace{1mm}
	
		\hspace{10.5mm} (1) $\omega_{l-1}^{(i)} \gets \operatorname{diag}({\phi'_{l}}^{(i)}) 
			W_{l} \omega_{l}^{(i)}, \forall i = 1,\ldots,T$;
		
		\hspace{10.5mm} (2) $W_{l} \gets W_{l} - \alpha \sum\limits_{i=1}^{T} \phi_{l-1}^{(i)} 
			(\omega_{l}^{(i)})^{\top}$ \vspace{1mm}}
	\Sc{Repeat from Step 1 until convergence \vspace{0mm}}
\end{algorithm}
The gradient of $J$ in the $l$-th weight matrix 
$W_{l} \in \mathbb{R}^{n_{l-1}\times n_{l}}$ with respect to the Euclidean metric 
can be computed as\
\begin{equation}
\label{eq:grad}
	\nabla_{\!\!J}(W_{l}) \!=\! \phi_{l-1} \big(\!
	\underbrace{ \Sigma'_{l} W_{l+1} \Sigma'_{l+1} \!\ldots\!
	W_{L} \Sigma'_{L}\!\nabla_{\!E}(\phi_{L})
	}_{=: \omega_{l} \in \mathbb{R}^{n_{l}}} \!\big)^{\!\top}\!\!\!,\!\!
\end{equation}
where $\Sigma'_{l} := \operatorname{diag}(\phi'_{l})$.
The gradient $\nabla_{\!\!J}(W_{l})$ is realised as a rank-one matrix, and
acts as a linear map $\nabla_{\!\!J}(W_{l}) \colon \mathbb{R}^{n_{l-1}\times n_{l}} 
\to \mathbb{R}$.
By exploring the layer-wise structure of the MLP, the corresponding vector $\omega_{l}$
can be computed recursively backwards from the output layer $L$, i.e.,
\begin{equation}
\label{eq:omega}
	\omega_{l} := \Sigma'_{l} W_{l+1} \omega_{l+1}, 
	\quad\text{for~all~}l = L-1,\ldots,1,
\end{equation}
with $\omega_{L} = \Sigma'_{L} \nabla_{\!E}(\phi_{L})$.
With such a backward mechanism in computing the gradient $\nabla_{\!J}(W_{l})$, we
recover the classic BP algorithm, see Algorithm~\ref{algo:01}. 
%

In what follows, we characterise critical points of the total loss function 
$\mathcal{J}$ as defined in Eq.~\eqref{eq:total_loss} by setting its gradient 
to zero, i.e., $\nabla_{\!\mathcal{J}}(\mathbf{W}) = 0$.
Explicitly, the gradient of $\mathcal{J}$ with respect to the $l$-th weight $W_{l}$
is computed by
\begin{equation}
\label{eq:cp}
	\nabla_{\!\mathcal{J}}(W_{l}) = \sum\limits_{i=1}^{T} 
	\phi_{l-1} \omega_{l}^{\top} \in \mathbb{R}^{n_{l-1}\times n_{l}},
\end{equation}
where $\phi_{l-1}$ and $\omega_{l}$ are respectively the $(l-1)$-th layer 
output and the $l$-th error vector as defined in Eq.~\eqref{eq:omega}. 
%
%
Simi\-lar to the recursive construction of the error vector $\omega_{l}$ as in 
Eq.~\eqref{eq:omega}, we construct a sequence of matrices as, for all 
$l = L-1,\ldots,1$, 
\begin{equation}
\label{eq:psi}
	\Psi_{l} := \Sigma'_{l} W_{l+1} \Psi_{l+1} \in 
	\mathbb{R}^{n_{l}\times n_{L}}, 
\end{equation}
with $\Psi_{L} = \Sigma'_{L} \in \mathbb{R}^{n_{L}\times n_{L}}$.
Then the vector form of the gradient $\nabla_{\!\mathcal{J}}(W_{l})$ can be 
written as 
\begin{equation}
\label{eq:gradient}
    \operatorname{vec} \big( \nabla_{\!\mathcal{J}}(W_{l}) \big) \!=\! 
    \sum\limits_{i=1}^{T} \big( \Psi_{l} \otimes \phi_{l-l} \big)
	\nabla_{\!E}(\phi_{L}), 
\end{equation}
where $\operatorname{vec}(\cdot)$ puts a matrix into the vector form, and 
$\otimes$ denotes the Kronecker product of matrices.
Then, by applying the previous calculation to all $T$ samples,
critical points of the total loss function $\mathcal{J}$ are characterised
as solutions of the following linear system in $\mathbf{W}$ 
\begin{equation}
\label{eq:tracy}
	\underbrace{
	\!\left[\!\!\! \begin{array}{ccc}
	\Psi_{L}^{(1)} \!\otimes\! \phi_{L-1}^{(1)} &
	\!\!\!\!\ldots\!\!\!\! &
	\Psi_{L}^{(T)} \!\otimes\! \phi_{L-l}^{(T)} \\
	\vdots & \!\!\!\!\ddots\!\!\!\! & \vdots \\
	\Psi_{1}^{(1)} \!\otimes\! \phi_{0}^{(1)} &
	\!\!\!\!\ldots\!\!\!\! &
	\Psi_{1}^{(T)} \!\otimes\! \phi_{0}^{(T)}
	\end{array}\!\!\!\right]\!
	}_{=: \mathbf{P}(\mathbf{W}) \in \mathbb{R}^{N_{net} \times (T n_{L})}}
	\,
	\underbrace{\!
	\left[\!\!\! \begin{array}{c}
	\nabla_{\!E}(\phi_{L}^{(1)}) \\
	\vdots \\
	\nabla_{\!E}(\phi_{L}^{(T)})
	\end{array}\!\!\!\right] \!
	}_{=: \bm{\varepsilon}(\mathbf{W}) \in \mathbb{R}^{T n_{L}}}
	\!=\! 0, \!
\end{equation}
where the superscript $(\cdot)^{(i)}$ indicates the corresponding term for
the $i$-th sample, and $\mathbf{P}(\mathbf{W})$ is the collection of the Jacobian matrices 
of the MLP for all $T$ samples.
Here, $N_{net}$ is the number of variables in the MLP, i.e.,
\begin{equation}
	N_{net} = \sum\limits_{l=1}^{L} n_{l-1} n_{l}. 
\end{equation}
The above parameterised linear equation system in $\bm{\varepsilon}(\mathbf{W})$ is strongly 
dependent on several factors, essentially all factors in designing an MLP, i.e., 
\emph{the MLP structure}, \emph{the activation function}, \emph{the error function}, 
\emph{given samples}, and \emph{the weight matrices}.
If the trivial solution $\bm{\varepsilon}(\mathbf{W}) = 0$ is reachable 
at some weights $\mathbf{W}^{*} \in \bm{\mathcal{W}}$, then a finite exact
approximator $\widehat{g}$ is realised by the corresponding MLP, i.e., 
$F(\mathbf{W}^{*}, \cdot) = \widehat{g}$.
Additionally, if the solution $\bm{\varepsilon} = 0$ is the only solution of the
linear equation system for all $\mathbf{W} \in \bm{\mathcal{W}}$, then any local minimum of
the loss function $\mathcal{J}$ is a global minimum.
Thus, we conclude the following theorem. 
\begin{theorem}[suboptimal local minima free condition]
\label{thm:local_free_general}
	Let an MLP architecture $\bm{\mathcal{F}}$ satisfy Assumption~\ref{ass:exist}
	for a specific learning task, and the error function $E$ satisfy 
	Principle~\ref{prin:error}.
	If the following two conditions are fulfilled for all $\mathbf{W} \in 
	\bm{\mathcal{W}}$, 
	\begin{enumerate}[(1)]
		\item the matrix $\mathbf{P}(\mathbf{W})$, as constructed in \eqref{eq:tracy}, is 
			\emph{non-zero},			
		\item the vector $\bm{\varepsilon}(\mathbf{W})$, as constructed in \eqref{eq:tracy},
			lies in the row span of $\mathbf{P}(\mathbf{W})$,
	\end{enumerate}
	then a finite exact approximator $\widehat{g}$ is realised at a global minimum 
	$\mathbf{W}^{*} \in \bm{\mathcal{W}}$, i.e., $F(\mathbf{W}^{*}, \cdot) = \widehat{g}$, 
	and the loss function $\mathcal{J}$ is free of suboptimal local minima, i.e.,
	any local minimum of $\mathcal{J}$ is a global minimum. 
\end{theorem}

Obviously, condition (1) in Theorem~\ref{thm:local_free_general} is quite easy to 
be ensured, while condition (2) is hardly possible to be realised, since it might 
require enormous efforts to design the space of MLPs $\bm{\mathcal{F}}$.
Nevertheless, if the rank of matrix $\mathbf{P}$ is equal to $T n_{L}$ for
all $\mathbf{W} \in \bm{\mathcal{W}}$, then the tri\-vial solution zero is the only 
solution of the parameterised linear system.
Hence, we have the following proposition as a special case of 
Theorem~\ref{thm:local_free_general}. 

\begin{proposition}[Strong suboptimal local minima free condition]
\label{prop:local_free_strong}
	Let an MLP architecture $\bm{\mathcal{F}}$ satisfy Assumption~\ref{ass:exist}
	for a specific learning task, and the error function $E$ satisfy 
	Principle~\ref{prin:error}.
	If the rank of matrix $\mathbf{P}(\mathbf{W})$ as constructed in \eqref{eq:tracy}
	is equal to $T n_{L}$ for all $\mathbf{W} \in \bm{\mathcal{W}}$, 
	%
	%
	then a finite exact approximator $\widehat{g}$ is realised at a global minimum 
	$\mathbf{W}^{*} \in \bm{\mathcal{W}}$, i.e., $F(\mathbf{W}^{*}, \cdot) = \widehat{g}$, 
	and the loss function $\mathcal{J}$ is 
	free of suboptimal local minima. 
\end{proposition}

Given the number of rows of $\mathbf{P}(\mathbf{W})$ being $N_{net}$, we suggest the second 
principle of ensuring performance of MLPs. 
\begin{principle}[Choice of the number of NN variables]
\label{prin:number}
	The total number of variables in an MLP architecture $N_{net}$
	needs to be greater than or equal to $T n_{L}$, i.e., 
	%
		$N_{net} \ge T n_{L}$. 
\end{principle}

In what follows, we investigate the possibility or difficulty to fulfil the condition,
i.e., $\operatorname{rank}\big(\mathbf{P}(\mathbf{W})\big) = T n_{L}$ for all 
$\mathbf{W} \in \bm{\mathcal{W}}$,
required in Proposition~\ref{prop:local_free_strong}.
%
%
Let us firstly construct the two identically partitioned matrices ($L \times T$ partitions),
by collecting the partitions $\Psi_{l}^{(i)}$'s and $\phi_{l}^{(i)}$'s, as 
\begin{equation}
\label{eq:bf_psi}
	\mathbf{\Psi} \!:=\!\! \left[\!\!\! \begin{array}{ccc}
	\Psi_{L}^{(1)} & \!\!\!\!\!\!\ldots\!\!\!\!\!\! & \Psi_{L}^{(T)} \\
	\!\!\!\vdots   & \!\!\!\!\!\!\ddots\!\!\!\!\!\! & \vdots \\
	\Psi_{1}^{(1)} & \!\!\!\!\!\!\ldots\!\!\!\!\!\! & \Psi_{1}^{(T)} 
	\end{array}\!\!\!\right]\!\!, 
	\quad\!
	~\text{and}~
	\mathbf{\Phi} \!:=\!\! \left[\!\!\! \begin{array}{ccc}
	\phi_{L-1}^{(1)} & \!\!\!\!\!\ldots\!\!\!\!\! & \phi_{L-1}^{(T)} \\
	\vdots           & \!\!\!\!\!\ddots\!\!\!\!\! & \vdots \\
	\phi_{0}^{(1)}   & \!\!\!\!\!\ldots\!\!\!\!\! & \phi_{0}^{(T)}
	\end{array}\!\!\!\right]\!.\!\!\! 
\end{equation}
%
%
Then, the matrix $\mathbf{P}(\mathbf{W})$ constructed on the left hand side of 
Eq.~\eqref{eq:tracy}
is computed as the \emph{Khatri-Rao product} of $\mathbf{\Psi}$ and $\mathbf{\Phi}$,
i.e., pairwise Kronecker products for all pairs of partitions in $\mathbf{\Psi}$ 
and $\mathbf{\Phi}$, denoted by 
\begin{equation}
\label{eq:jacobi}
	\mathbf{P}(\mathbf{W}) := \mathbf{\Psi} \odot \mathbf{\Phi}. 
\end{equation}
%
%
Each row block of $\mathbf{P}(\mathbf{W})$ associated with a specific
layer $l$ is by construction the \emph{Khatri-Rao product} of the corresponding 
row blocks in $\mathbf{\Psi}$ and $\mathbf{\Phi}$, i.e., 
$\mathbf{\Psi}_{l} \odot \mathbf{\Phi}_{l-1}$ with 
$\mathbf{\Psi}_{l} := \big[\Psi_{l}^{(1)}, \ldots, \Psi_{l}^{(T)}\big]$ and
$\mathbf{\Phi}_{l-1} := \big[\phi_{l-1}^{(1)}, \ldots, \phi_{l-1}^{(T)}\big]$.
We firstly investigate the rank property of row blocks of $\mathbf{P}(\mathbf{W})$
in the following lemma\footnote{The proof is given in the provided supplements.}.

\begin{proposition}
\label{prop:row_rank}
	Given a collection of matrices $\Psi_{i} \in \mathbb{R}^{n_{l} \times n_{L}}$ 
	and a collection of vectors $\phi_{i} \in \mathbb{R}^{n_{l-1}}$, 
	for $i = 1, \ldots, T$, let $\mathbf{\Psi} := [ \Psi_{1}, \ldots \Psi_{T} ]
	\in \mathbb{R}^{n_{l} \times (n_{L} T)}$
	and $\mathbf{\Phi} = [ \phi_{1}, \ldots, \phi_{T}] \in \mathbb{R}^{n_{l-1} \times T}$.
	Then the rank of the Khatri-Rao product 
	$\mathbf{\Psi} \odot \mathbf{\Phi}$ is bounded from below by 
	\begin{equation}
		\operatorname{rank}(\mathbf{\Psi} \odot \mathbf{\Phi})
		\ge
		n_{l} \operatorname{rank}(\mathbf{\Phi}) + 
		\sum\limits_{i=1}^{T}\operatorname{rank}(\Psi_{i}) - T n_{L}. 
	\end{equation}
	If all matrices $\Psi_{i}$'s and $\mathbf{\Phi}$ are of full rank, then the rank of 
	$\mathbf{\Psi} \odot \mathbf{\Phi}$ has the following properties: 
	\begin{enumerate}[(1)]
		\item If $n_{l} \le n_{L}$, then 
			$\operatorname{rank}(\mathbf{\Psi} \odot \mathbf{\Phi}) \ge n_{l} 
			\operatorname{rank}(\mathbf{\Phi})$;
		\item If $n_{l} > n_{L}$ and $n_{l-1} \ge T$, then 
			$\operatorname{rank}(\mathbf{\Psi} \odot \mathbf{\Phi}) \ge T n_{L}$;
		\item If $n_{l} > n_{L}$ and $n_{l-1} < T$, then
    		$\operatorname{rank}(\mathbf{\Psi} \odot \mathbf{\Phi}) \ge n_{L}$.
	\end{enumerate}
\end{proposition}

Unfortunately, stacking these row blocks $\mathbf{\Psi}_{l} \odot \mathbf{\Phi}_{l-1}$ 
for $l=1, \ldots, L$ together to construct $\mathbf{P}(\mathbf{W})$ cannot bring better 
knowledge about the rank of $\mathbf{P}(\mathbf{W})$. 

\begin{proposition}
\label{prop:rank_jacobi}
	For an MLP architecture $\bm{\mathcal{F}}$, the rank of $\mathbf{P}(\mathbf{W})$ 
	as defined in Eq.~\eqref{eq:jacobi} is bounded from below by 
	\begin{equation}
	\begin{split}
		\operatorname{rank}\!\big(\mathbf{P}(\mathbf{W})\big) 
		\ge & \sum\limits_{l=1}^{L} n_{l} \operatorname{rank}\big(\mathbf{\Phi}_{l-1}\big) 
		+ \sum\limits_{l=1}^{L} \sum\limits_{i=1}^{T} 
		\operatorname{rank}\big(\Psi_{l}^{(i)}\big) ~- \\
		& - \sum\limits_{l=1}^{L-1} T n_{l} - L T n_{L}.
	\end{split}
	\end{equation}
\end{proposition}

Clearly, in order to have the rank of $\mathbf{P}(\mathbf{W})$
better bounded from below, it is important to ensure higher rank of all
$\Psi_{l}^{(i)}$'s and $\mathbf{\Phi}_{l-1}$'s. 
By choosing appropriate  
activation functions in hidden layers, all $\mathbf{\Phi}_{l-1}$'s
can have full rank \cite{itoy:acm96,shah:tnn99}.
Then, in what follows, we present three heuristics aiming to keep
all $\Psi_{l}^{(i)}$'s being of full rank as much as possi\-ble.
%
%
By the construction of $\Psi_{l}^{(i)}$ as specified in Eq.~\eqref{eq:psi}, 
i.e., matrix product of $\Sigma_{l}^{'(i)}$'s and $W_{l}^{(i)}$'s,
it is reasonable to ensure full rankness of all $\Sigma_{l}^{'(i)}$'s 
and $W_{l}^{(i)}$'s.
\begin{principle}[Constraints on MLP weights]
\label{prin:weight}
	All weight matrices $W_{l} \in \mathbb{R}^{n_{l-1} \times n_{l}}$  for all 
	$l = 1, \ldots, L$ are of full rank.
\end{principle}

\begin{remark}
	Note that, the full rank constraint on the weight matrices does not introduce 
	new local minima of the constrained total loss function.
	However, whether such a constraint may exclude all global minima of the unconstrained 
	total loss function for a given MLP architecture is still an open problem.
\end{remark}

\begin{principle}[Choice of activation functions]
\label{prin:act}
	The derivative of activation function $\sigma$ is non-zero for all
	$z \in \mathbb{R}$, i.e., $\sigma'(z) \neq 0$.
\end{principle}

\begin{remark}
	It is trivial to verify that most of popular differentiable activation functions, 
	such as the \emph{Sigmoid}, $\operatorname{tanh}$, SoftPlus, and 
	SoftSign, satisfy Principle~\ref{prin:act}. 
	Note, that the Identity activation function, which is employed in 
	the output layer, also satisfies this principle. 
	However, potentially vanishing gradient of squashing activation functions can 
	be still an issue in practice due to finite machine precision. 
	Therefore, activation functions without vanishing gradients, e.g. the 
	Bent identity or the leaky ReLU (non-differentiable at the origin), might be
	preferred. 
\end{remark}

However, even when both Principles \ref{prin:weight} and \ref{prin:act} are fulfilled, 
matrices $\Psi_{l}^{(i)}$'s still cannot be guaranteed to have full rank, according to 
the Sylvester's rank inequality \cite{simo:book12}.
Hence, we need to prevent loss of rank in each $\Psi_{l}^{(i)}$ due to 
matrix product, i.e., to preserve the smaller rank of two matrices after a 
matrix product.

\begin{principle}[Choice of the number of hidden units]
\label{prin:hidden}
	Given an MLP with hidden layers, the numbers
	of units in three adjacent layers, namely $n_{l-1}$, $n_{l}$, and $n_{l+1}$, 
	satisfy the following inequality with $l \le L-2$
	\begin{equation}
	\label{eq:hidden}
		n_{l} \le \max\{n_{l-1}, n_{l+1}\}. 
	\end{equation}
\end{principle}

\begin{remark}
	The condition $l \le L-2$ together with $l-1 \ge 0$ implies $L \ge 3$, 
	i.e., the inequality in Eq.~\eqref{eq:hidden} takes effect when there is more
	than one hidden layer,
	since Principle \ref{prin:weight} and \ref{prin:act} are sufficient to ensure 
	$\Psi_{L}^{(i)}$'s and $\Psi_{L-1}^{(i)}$'s to have full rank.
	However, for $l \le L-2$, the inequality 
	in Eq.~\eqref{eq:hidden} together with Principle \ref{prin:weight} and 
	\ref{prin:act} ensures no loss of rank in all $\Psi_{l}^{(i)}$, i.e.,
		$\operatorname{rank}\!\big(\Psi_{l}^{(i)}\big) = \min \{ n_{l}, \ldots, n_{L} \}$.
\end{remark}

\section{Hessian analysis of MLP training}
\label{sec:04}
It is well known that gradient descent algorithms can suffer from slow
convergence.
Information from the Hessian matrix of the loss function is critical 
for developing efficient numerical algorithms.
Specifically, definiteness of the Hessian matrix is an indicator to the 
isolatedness of the criti\-cal points, which will affect significantly the 
convergence speed of the corresponding algorithm.

We start with the Hessian of the \emph{sample-wise MLP loss function} $J$
as defined in Eq.~\eqref{eq:sample_loss}, which is a bilinear operator 
$\mathsf{H}_{J} \colon \mathbb{R}^{N_{net}} \times \mathbb{R}^{N_{net}} \to
\mathbb{R}$, computed by the second directional derivative of $J$.
For the sake of readability, we only present one component of the second directional 
derivative with respect to two specific layer indices $k$ and $l$, i.e., 
\begin{equation}
\label{eq:dt2}
\begin{split}
	& \operatorname{D}^{2} \!J(\mathbf{W})(H_{k},H_{l}) = \tfrac{\dt^{2}}{\dt t^{2}}
	J_{l,k}(\mathbf{W} + t \mathbf{H}) {\big|}_{t=0} \\
	= & \operatorname{D}^{2} E(\phi_{L}) \big( \!\operatorname{D}\!F(W_{l}) \!\cdot\! H_{l}, 
	\operatorname{D}\!F(W_{k}) \!\cdot\! H_{k} \big) ~+ \\
	& + \!\sum\limits_{i=1}^{L} \sum\limits_{l,k>i}^{L} \!
	\left( W_{i}^{\top}\!\ldots \Sigma'_{k} H_{k}^{\top}
	\phi_{k-1} \right)^{\!\top}\!\!\diag\!\big(\nabla_{\!J}(W_{i})\big) 
	\Sigma''_{i} 
	\left( W_{i}^{\top} \ldots \Sigma'_{l} H_{l}^{\top} 
	\phi_{l-1} \right),
\end{split}
\end{equation}
where $\Sigma''_{i} \in \mathbb{R}^{n_{i} \times n_{i}}$ is the diagonal matrix with
its diagonal entries being the second derivative of the activation functions.
Since exact learning of finite samples is assumed at a global minimum $\mathbf{W}^{*}$,
gradients of the error function at all samples are simply zero, i.e., 
$\nabla_{\!E}(\phi_{L}^{*(i)}) = 0$ for all $i = 1, \ldots, T$. 
Consequently, the second summand in the last equation in Eq.~\eqref{eq:dt2} vanishes,
as $\nabla_{\!J}(W_{i}) = 0$ for all $i = 1, \ldots, L$.
Then, the Hessian $\mathsf{H}_{J}(\mathbf{W})$ evaluated at $\mathbf{W}^{*}$ in
direction $\mathbf{H} \in \bm{\mathcal{W}}$ is given as
\begin{equation}
\begin{split}
	\operatorname{D}^{2}\! J(\mathbf{W}^{*})(\mathbf{H},\mathbf{H}) = &~
	\mathsf{H}_{J}(\mathbf{W}^{*})(\mathbf{H},\mathbf{H}) \\
	= & \!\sum\limits_{l,k=1}^{L} \!\!\big(\!\operatorname{D}\!
	F(W_{l}^{*}) \!\cdot\! H_{l}\big)^{\!\top}
	\mathsf{H}_{E}(\phi_{L}^{*}) 
	\big(\!\operatorname{D}\!F(W_{k}^{*}) \!\cdot\! H_{k} \big),
\end{split}
\end{equation}
where $\mathsf{H}_{E}(\phi_{L}) \colon \mathbb{R}^{n_{L}} \times \mathbb{R}^{n_{L}}
\to \mathbb{R}$ is the Hessian of the error function $E$ with respect to
the output of the MLP $\phi_{L}$.
By a direct computation, we have the Hessian of the total loss function 
$\mathcal{J}$ at a global minimum $\mathbf{W}^{*}$ in a matrix form as 
\begin{equation}
\label{eq:ground_hess}
	\mathsf{H}_{\mathcal{J}}(\mathbf{W}^{*}) = 
	\mathbf{P}(\mathbf{W}^{*})~\!
	\mathbf{H}_{E}(\mathbf{W}^{*})
	\big(\mathbf{P}(\mathbf{W}^{*})\big)^{\top}, 
\end{equation}
where $\mathbf{P}(\mathbf{W}^{*})$ is the Jacobian matrix of the MLP
evaluated at $\mathbf{W}^{*}$ as defined in Eq.~\eqref{eq:tracy},
and $\mathbf{H}_{E}(\mathbf{W}^{*}) := \diag\big(\mathsf{H}_{E}(\phi_{L}^{*(1)}), 
\ldots, \mathsf{H}_{E}(\phi_{L}^{*(T)}) \big) \in \mathbb{R}^{Tn_{L} \times Tn_{L}}$
is a block diagonal matrix of all Hessians of $E$ for all $T$ samples
evaluated at $\mathbf{W}^{*}$.
It is then trivial to conclude the following proposition about the rank of 
$\mathsf{H}_{\mathcal{J}}(\mathbf{W}^{*})$.

\begin{proposition}
	The rank of the Hessian of the total loss fun\-ction $\mathcal{J}$ at
	a global minimum $\mathbf{W}^{*}$ is bounded from above by
	\begin{equation}
		\operatorname{rank}(\mathsf{H}_{\mathcal{J}}(\mathbf{W}^{*})) 
		\le T n_{L}.
	\end{equation}
\end{proposition}

\begin{remark}
\label{rmk:degenerate}
	When Principle~\ref{prin:number} is assumed, it is easy to see 
	\begin{equation}
		\operatorname{rank}(\mathsf{H}_{J}(\mathbf{W}^{*})) \le T n_{L}
		\le N_{net}.
	\end{equation}
	Namely, when an MLP is designed from scratch without insightful knowledge
	and reaches exact learning, then it is very likely that the Hessian is 
	degenerate, i.e., the classic BP algorithm will suffer significantly from 
	slow convergence.
	Moreover, it is also obvious that conditions constructed in the main result 
	of \cite{nguy:icml17} (Theorem 3.8), i.e., the existence of non-degenerate 
	global minima, is hardly possible to be satisfied for the setting considered
	in \cite{nguy:icml17}.
\end{remark}

If Proposition~\ref{prop:local_free_strong} holds true, i.e.,
$\operatorname{rank}\big(\mathbf{P}(\mathbf{W}^{*})\big) = T n_{L}$,
then the rank of $\mathsf{H}_{\mathcal{J}}(\mathbf{W}^{*})$ will depend on
the rank of $\mathbf{H}_{E}(\mathbf{W}^{*})$.
To ensure $\mathbf{H}_{E}(\mathbf{W}^{*})$ to have full rank, 
we need to have a non-degenerate Hessian for the error function $E$
at global minima, i.e., $E$ is a Morse function \cite{miln:book63}. 
Hence, in addition to Principle~\ref{prin:error}, we state the following
principle on the choice of error functions.

\begin{customthm}{1.a}[Strong choice of error function]
\label{prin:error2}
	In addition to Principle~\ref{prin:error}, the error function $E$ is 
	Morse, i.e., the Hessian $\mathsf{H}_{E}(\phi_{L})$
	is non-degenerate for all $\phi_{L} \in \mathbb{R}^{n_{L}}$.
\end{customthm}

\section{Case studies}
\label{sec:05}
In this section, we evaluate our results in the previous sections
by two case studies, namely loss surface analysis of training MLPs 
with one hidden layer, and development of an approximate Newton's algorithm.

\subsection{MLPs with one hidden layer}
\label{sec:51}
Firstly, we revisit some classic results on MLPs with only one hidden 
layer \cite{yuxi:tnn92,yuxi:tnn95}, and 
exemplify our analysis using the classic XOR problem
\cite{lisb:network91,hame:nn98,spri:amai99,spri:tnn99}.
For a learning task with $T$ unique training samples, a finite exact approximator 
is realisable with a two-layer MLP ($L=2$) having $T$ units in the hidden layer, and
its training process exempts from suboptimal local minima. 

\begin{proposition}
\label{prop:hidden_t}
	Let an MLP architecture with one hidden layer satisfy
	Principle~\ref{prin:error}, \ref{prin:weight}, and \ref{prin:act}.
	Then, for a learning task with $T$ unique training samples, 
	if the following two conditions are fulfilled:
	\begin{enumerate}[(1)]
		\item There are $T$ units in the hidden layer, i.e., $n_{1} = T$, 
		\item $T$ unique samples produce a basis in the output space of 
			the hidden layer for all $W_{1} \in \mathbb{R}^{n_{0} \times n_{1}}$,
	\end{enumerate}
	then a finite exact approximator $\widehat{g}$ is realised at a global minimum 
	$\mathbf{W}^{*} \in \bm{\mathcal{W}}$, i.e., $F(\mathbf{W}^{*}, \cdot) = \widehat{g}$, 
	and the loss function $\mathcal{J}$ is free of suboptimal local minima. 
\end{proposition}

When the scalar-valued bias $b_{l,k}$ in each unit map, as 
in Eq.~\eqref{eq:unit_map}, is set to be a free variable, a dummy unit 
is introduced in each layer, except the output layer. The dummy unit always feeds 
a constant input of one to its successor layer.
Results in \cite{yuxi:tnn92,yuxi:tnn95} claim that, with the presence of 
dummy units, only $T-1$ units in the hidden layer are required to achieve exact 
learning and eliminate all suboptimal local minima.
Such a statement has been shown to be false by counterexample utilising the XOR problem
\cite{spri:tnn99}.
In the following proposition, we reinvestigate this problem as a concrete example of 
applying Proposition~\ref{prop:rank_jacobi}.

\begin{proposition}
\label{prop:hidden_s} 
	Let a two-layer MLP architecture $\bm{\mathcal{F}}(n_{0},n_{1},n_{2})$ with 
	dummy units and
	$n_{2} \le n_{1} \le T$ satisfy Principle~\ref{prin:error}, \ref{prin:weight}, and 
	\ref{prin:act}, and $\mathbf{1} := [1, \ldots, 1]^{\top} \in \mathbb{R}^{T}$.
	For a learning task with $T$ unique samples $X \in \mathbb{R}^{n_{0} \times T}$, 
	we have
	\begin{enumerate}[(1)]
		\item if $\operatorname{rank}([X^{\top}, \mathbf{1}]) = n_{0}$, then
			\begin{equation}
				\!\!\!\!\!\operatorname{rank}\!\big(\mathbf{P}(\mathbf{W})\big)
				\ge \max \big\{ n_{1}n_{2}, n_{1}(n_{0}+n_{2}-T) \big\}; 
			\end{equation}
		\item if $\operatorname{rank}([X^{\top}, \mathbf{1}]) = n_{0}+1$, then 
			\begin{equation}
				\!\!\!\!\!\operatorname{rank}\!\big(\mathbf{P}(\mathbf{W})\big)
				\!\ge\! \max \!\big\{ n_{1}n_{2}, n_{1}(n_{0}+n_{2}-T+1) \big\}. 
			\end{equation}
	\end{enumerate}
\end{proposition}

In the rest of this section, we examine these statements on the XOR problem
\cite{hame:nn98,spri:amai99,spri:tnn99}.
%
%
Two specific MLP structures have been extensively studied in the literature,
namely, the $\bm{\mathcal{F}}(2,2,1)$ network and 
the $\bm{\mathcal{F}}(2,3,1)$ network.
Squashing activation functions are used in the hidden layer. 
Dummy units are introduced in both the input and hidden layer.
The XOR problem has $T = 4$ unique input samples 
\begin{equation}
	X \!:=\! \begin{bmatrix}
	[\!\!\!\!\!\!&x_{1} \!\!\!\!\! & x_{2} \!\!\!\!\! & x_{3} \!\!\!\!\! & x_{4}
	& \!\!\!\!\!] \\
	[\!\!\!\!\!\!&1     \!\!\!\!\! & 1     \!\!\!\!\! & 1     \!\!\!\!\! & 1
	& \!\!\!\!\!]
	\end{bmatrix}
	\!=\! \left[
	\begin{matrix}
		0 & 0 & 1 & 1 \\
		0 & 1 & 0 & 1 \\
		1 & 1 & 1 & 1
	\end{matrix}
	\right] \!\in\! \mathbb{R}^{3 \times 4}, 
\end{equation}
where the last row is due to the dummy unit in the input layer and 
$\operatorname{rank}(X) = 3$.
Their associated desired outputs are specified as 
\begin{equation}
	Y := [y_{1}~y_{2}~y_{3}~y_{4}] = [0, 1, 1, 0] \in \mathbb{R}^{1 \times 4}.
\end{equation}
Then a squared loss function is defined as
\begin{equation}
	\mathcal{J}_{xor}(\mathbf{W}) := \frac{1}{4}
	\sum\limits_{i=1}^{4} \left( F(\mathbf{W}, x_{i}) - y_{i} \right)^{2},
\end{equation}
where $\mathbf{W} := \{W_{1}, W_{2}\} \in \mathbb{R}^{3 \times n_{1}} \times 
\mathbb{R}^{(n_{1}+1) \times 1}$.
In order to identify local minima of the loss function $\mathcal{J}_{xor}$,
we firstly compute $n_{1}(n_{0}+n_{2}-T+1) = 0$, 
according to situation (2) in Proposition~\ref{prop:hidden_s}.
Hence, we need to investigate the rank of the Jacobian matrix 
\begin{equation}
\label{eq:xor}
	\mathbf{P}_{xor}(\mathbf{W}) := \left[\!\!\! \begin{array}{ccc}
	1 \otimes \phi_{1}^{(1)} &
	\!\!\ldots\!\! &
	1 \otimes \phi_{1}^{(4)} \\[1mm]
	\Psi_{1}^{(1)} \otimes x_{1} &
	\!\!\ldots\!\! &
	\Psi_{1}^{(4)} \otimes x_{4}
	\end{array}\!\!\!\right].
\end{equation}

Firstly, let us have a look at the $\bm{\mathcal{F}}(2,2,1)$ XOR network.
By considering the dummy unit in the hidden layer, the matrix 
$\Phi_{1} := [\phi_{1}^{(1)}, \ldots, \phi_{1}^{(4)}] \in \mathbb{R}^{3 \times 4}$ 
has the last row as $[1,1,1,1]$.
Then, according to Proposition~\ref{prop:row_rank}, the first row block in 
$\mathbf{P}_{xor}$ in Eq.~\eqref{eq:xor} has the smallest rank of two, while 
the second row block has the smallest rank of one.
Hence, the rank of $\mathbf{P}_{xor}(\mathbf{W})$ is lower bounded by two, and
local minima can exist for training the $\bm{\mathcal{F}}(2,2,1)$ XOR network 
\cite{spri:amai99,lisb:network91}.

Similarly, for the $\bm{\mathcal{F}}(2,3,1)$ network, i.e., $n_{1} = 3$, 
although the first row block in $\mathbf{P}_{xor}$ has potentially the 
largest rank of four, it is still not immune from collapsing to lower rank of three.
Meanwhile, the second row block in $\mathbf{P}_{xor}$ has also the smallest rank of one.
Therefore, the rank of $\mathbf{P}_{xor}(\mathbf{W})$ is lower bounded by three. 
As a sequel, there still exist undesired local minima in training the $\bm{\mathcal{F}}(2,3,1)$ 
XOR network \cite{spri:tnn99}.

\subsection{An approximate Netwon's algorithm} 
\label{sec:52}
It is important to notice that the Hessian $\mathsf{H}_{\mathcal{J}}(\mathbf{W}^{*})$ 
is neither diagonal nor block diagonal, which differs from the 
exist\-ing approximate strategies of the Hessian in \cite{batt:nc92,wang:nn98}.
A classic Newton's method for minimising the total loss function 
$\mathcal{J}$ requires to compute the exact Hessian of $\mathcal{J}$ from its 
second directional derivative as in Eq.~\eqref{eq:dt2}.
The complexity of computing the second summand on the right hand side of 
the last equality in Eq.~\eqref{eq:dt2} is of order $O(L^{3})$ in the number of layers $L$.
Namely, an implementation of exact Newton's method becomes much more 
expensive, when an MLP gets deeper.
Motivated by the fact that this computationally expensive term vanishes at a global 
mini\-mum, shown in Eq.~\eqref{eq:ground_hess}, we propose to approximate the Hessian 
of $\mathcal{J}$ at an arbitrary weight $\mathbf{W}$ with the following expression
\begin{equation}
\label{eq:approx_hess}
	\widetilde{\mathsf{H}}_{\mathcal{J}}(\mathbf{W}) = 
	\mathbf{P}(\mathbf{W})~\!
	\mathbf{H}_{E}(\mathbf{W})
	\big(\mathbf{P}(\mathbf{W})\big)^{\top}, 
\end{equation}
where $\mathbf{H}_{E}(\mathbf{W}) := \diag\big(\mathsf{H}_{E}(\phi_{L}^{(1)}), 
\ldots, \mathsf{H}_{E}(\phi_{L}^{(T)}) \big)$, and $\mathbf{P}(\mathbf{W})$ is 
the Jacobian matrix of the MLP as defined in Eq.~\eqref{eq:tracy}.
With this approximation, we can construct an approximate Newton's algorithm
to minimise the total loss function $\mathcal{J}$.
Specifically, for the $k$-th iterate $\mathbf{W}^{(k)}$, an approximate Newton's direction 
is computed by solving the following linear system for 
$\bm{\xi}_{N}^{(k)} \in \mathbb{R}^{N_{net}}$ 
\begin{equation}
\label{eq:newton}
	\widetilde{\mathsf{H}}_{\mathcal{J}}(\mathbf{W}^{(k)}) \bm{\xi}_{N}^{(k)} = 
	\operatorname{vec} \big( \nabla_{\!\mathcal{J}}(\mathbf{W}^{(k)}) \big),
\end{equation}
where $\nabla_{\!\mathcal{J}}(\mathbf{W}^{(k)})$ is the gradient of $\mathcal{J}$
at $\mathbf{W}^{(k)}$.
When Principle~\ref{prin:error2} holds, the approximate Newton's direction
$\bm{\xi}_{N}^{(k)}$ can be computed as $\bm{\xi}_{N}^{(k)} = 
\mathbf{P}(\mathbf{W}^{(k)}) \bm{\xi}^{(k)}$ with $\bm{\xi}^{(k)} \in \mathbb{R}^{T n_{L}}$ 
solving the following linear system
\begin{equation}
	\widetilde{\mathsf{H}}_{\mathcal{J}}(\mathbf{W}^{(k)}) \mathbf{P}(\mathbf{W}^{(k)}) 
	\bm{\xi}^{(k)} = \mathbf{P}(\mathbf{W}^{(k)}) \bm{\varepsilon}(\mathbf{W}^{(k)}).
\end{equation}
Then the corresponding Newton's update is defined as
\begin{equation}
\label{eq:approx_newton}
	\mathbf{W}^{(k+1)} = \mathbf{W}^{(k)} - \alpha~\mathbf{P}(\mathbf{W}^{(k)}) \bm{\xi}^{(k)},
\end{equation}
where $\alpha > 0$ is a suitable step size.
\begin{algorithm}[t!]
\caption{An approximate Newton's algorithm for supervised learning (batch learning).} 
\label{algo:newtonbp} 
\SetAlgoNoLine
	\SetKwHangingKw{IN}{Input~:}
	\SetKwHangingKw{Sa}{Step~1:}
	\SetKwHangingKw{Sb}{Step~2:}
	\SetKwHangingKw{Sc}{Step~3:}
	\SetKwHangingKw{Sd}{Step~4:}
	\SetKwHangingKw{Se}{Step~5:}
	\SetKwHangingKw{Sf}{Step~6:}
	\SetKwHangingKw{Sg}{Step~7:}
	\SetKwHangingKw{OUT}{\hspace{-1.6mm}Output:}

	\IN{Samples $(x_{i},y_{i})_{i=1}^{T}$, an MLP architecture 
		$\mathcal{F}(n_{1},\ldots,n_{L})$,
		and initial weights $\mathbf{W}$
		\vspace{1mm}}
		
	\OUT{Accumulation point $\mathbf{W}^{*}$ \vspace{1mm}}
				
	\Sa{For $i = 1,\ldots,T$, feed samples $x_{i}$ through the 
		MLP to compute $\phi_{l}^{(i)}\!$ and ${\phi'_{l}}^{(i)}\!$ for $l = 1, \ldots, L$ 
		\vspace{1mm}}
	
	\Sb{Compute 
	
	\hspace{10.5mm} (1) $\Psi_{L}^{(i)} := \operatorname{diag}({\phi'_{L}}^{(i)})
		\in \mathbb{R}^{n_{L}\times n_{L}}$; \vspace{1mm}
	
	\hspace{10.5mm} (2) $\omega_{L}^{(i)} := \operatorname{diag}({\phi'_{L}}^{(i)}) 
		\nabla_{\!E}(\phi_{L}^{(i)}) \in \mathbb{R}^{n_{L}}$; \vspace{1mm}
	
	\hspace{10.5mm} (3) $\mathsf{H}_{E}(\phi_{L}^{(i)}) \in \mathbb{R}^{n_{L}\times n_{L}}$ 
	\vspace{1mm}}
	
	\Sc{For $l = L,\ldots,1$ and $i = 1,\ldots,T$, compute \vspace{1mm}
	
		\hspace{10.5mm} (1) $\Psi_{l-1}^{(i)} := \operatorname{diag}({\phi'_{l-1}}^{(i)})
			W_{l} \Psi_{l}^{(i)} \in \mathbb{R}^{n_{l-1}\times n_{l}}$; \vspace{1mm}
		
		\hspace{10.5mm} (2) $\omega_{l-1}^{(i)} \gets \Psi_{l-1}^{(i)} 
			\nabla_{\!E}(\phi_{l}^{(i)}) \in \mathbb{R}^{n_{l-1}}$
		\vspace{1mm}}
	\Sd{For $l = L,\ldots,1$, compute the gradient \\
		\hspace{10.5mm} $\nabla_{\!\mathcal{J}}(W_{l}) = \sum\limits_{i=1}^{T} \phi_{l-1}^{(i)} 
			(\omega_{l}^{(i)})^{\top}$
	\vspace{1mm}}

	\Se{Compute the approximate Hessian 
		$\widetilde{\mathsf{H}}_{\mathcal{J}}(\mathbf{W})$ according to 
		Eq.~\eqref{eq:approx_hess}
		\vspace{1mm}}
			
	\Sf{Compute the Newton update $\mathbf{W} \gets \mathbf{W} - \alpha~
			\mathbf{P}(\mathbf{W}) \bm{\xi}$, where $\bm{\xi}$ solves \vspace{1mm}
			
			\hspace{10.5mm} $\widetilde{\mathsf{H}}_{\mathcal{J}}(\mathbf{W}) 
			\mathbf{P}(\mathbf{W}) \bm{\xi} = \mathbf{P}(\mathbf{W}) \bm{\varepsilon}
			(\mathbf{W})$\vspace{1mm}}

	\Sg{Repeat from Step 1 until convergence\vspace{1mm}}
\end{algorithm}

%
\begin{figure*}[!t]
	\centering
	\subfigure[\emph{Convergence with exact learning}]{
		\label{fig:quad}
		\includegraphics[width=0.48\textwidth]
		{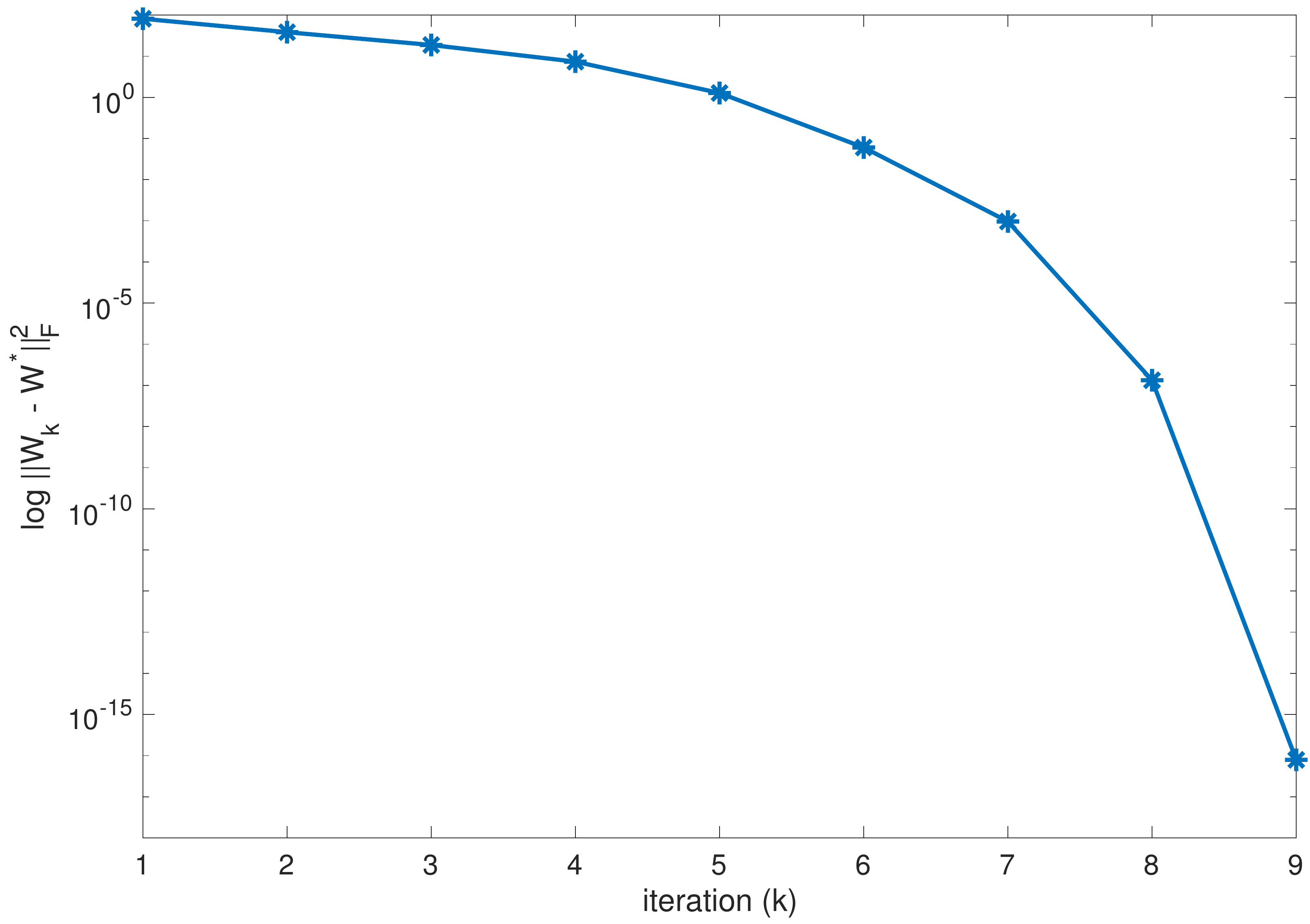}
	}
	\subfigure[\emph{Impact of vanishing gradient}]{
		\label{fig:vanish}
		\includegraphics[width=0.48\textwidth]
		{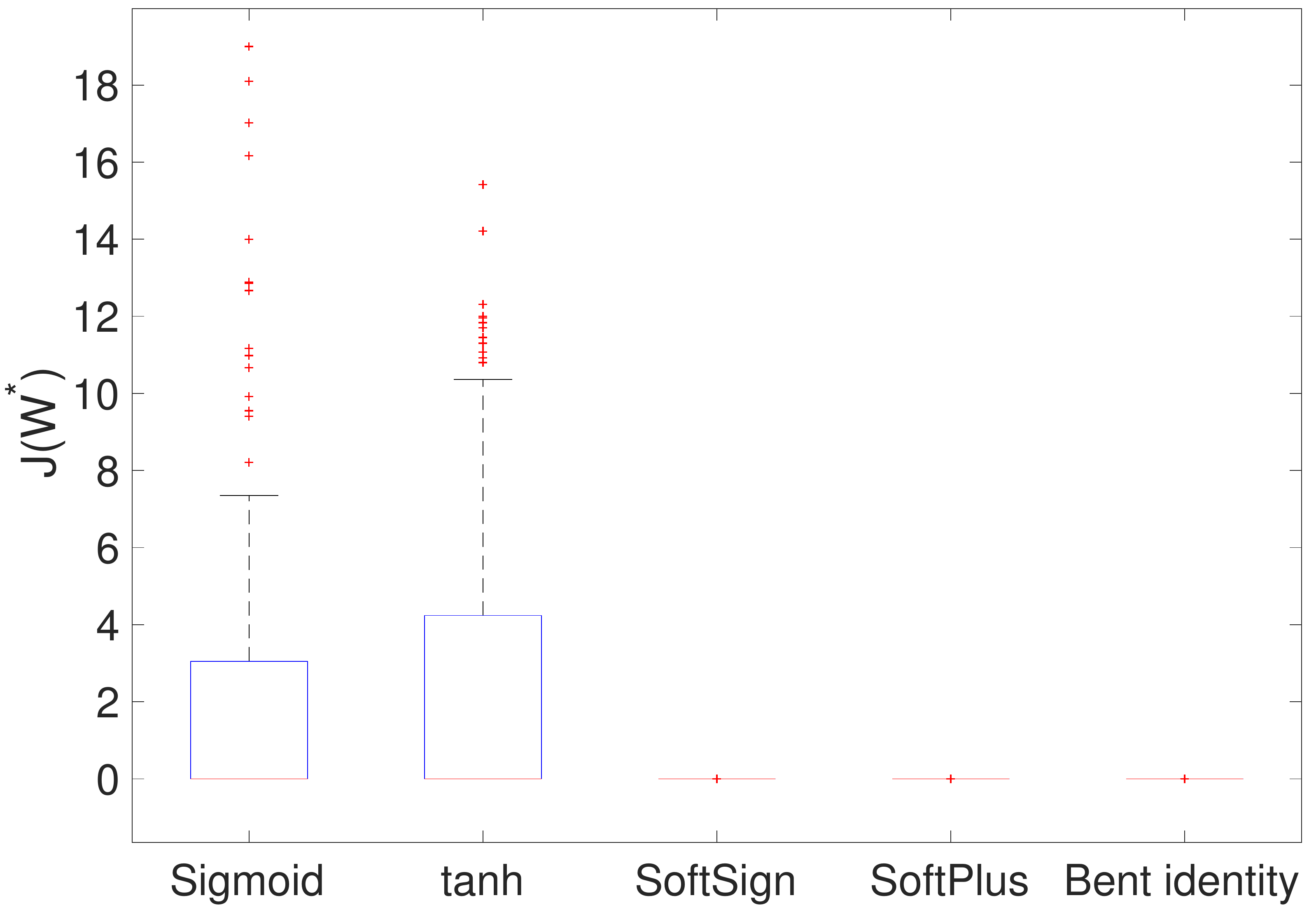}
	}
	\subfigure[\emph{Convergence without exact learning}]{
		\label{fig:conv}
		\includegraphics[width=0.48\textwidth]
		{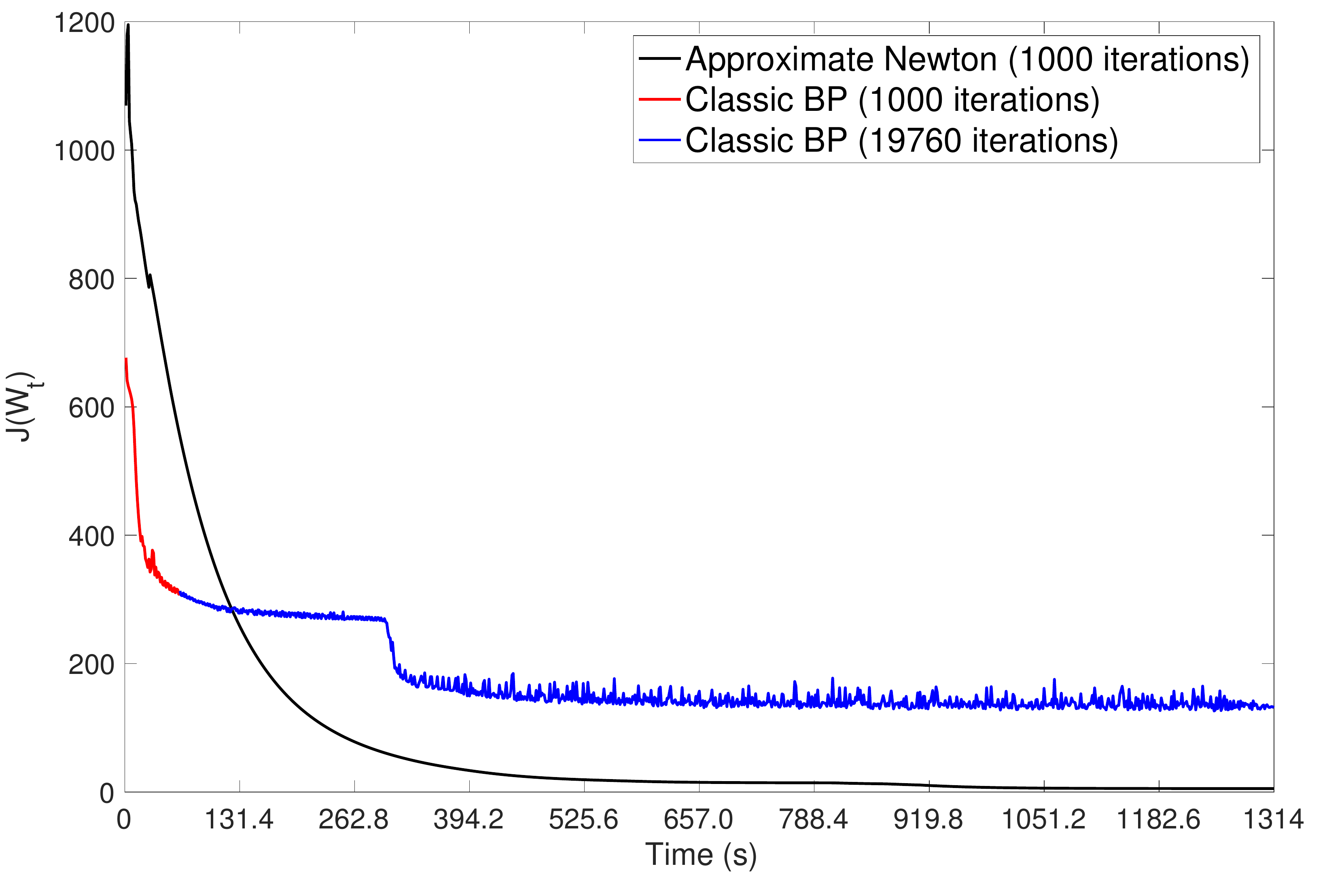}
	}
	\vspace{2mm}
	\caption{Investigation of an approximate Newton's algorithm, 
		a.k.a. the Generalised Gauss-Newton algorithm.}
	\label{fig:4r}
\end{figure*}

\begin{remark}
	The approximate Hessian proposed in Eq.~\eqref{eq:approx_hess} is by construction 
	positive semi-definite at arbitrary weights, while definiteness of the exact 
	Hessian is not conclusive.
	It is trivial to see that the approximate Hessian coincides with the ground-truth 
	Hessian as Eq.~\eqref{eq:ground_hess} at global minima.
	Hence, when $\alpha = 1$, the corresponding approximate Newton's algorithm induced 
	by the update rule in Eq.~\eqref{eq:approx_newton} shares the same local quadratic 
	convergence properties to a global minimum as the exact Newton's method 
	(see Section~\ref{sec:06}).
	
	In general, computing the approximate Newton's update as defined in 
	Eq.~\eqref{eq:newton} can be computationally expensive.
	Interestingly, the approximate Newton's algorithm is indeed the state of 
	the art GGN algorithm developed in \cite{schr:nc02}.
	Efficient implementations of the GGN algorithm have been extensively 
	explored in \cite{mart:icml10,fair:nc12,bote:icml17}. 
	In the next section, we investigate the theoretical convergence properties 
	of the approximate Newton's algorithm, i.e., the GGN algorithm. 
\end{remark}


\section{Numerical evaluation} 
\label{sec:06}
%
In this section, we investigate performance of the Approximate
Newton's (AN) algorithm, i.e., the GGN algorithm.
We test the algorithm on the four regions classification benchmark, as 
originally proposed in \cite{sing:nips89}.
In $\mathbb{R}^{2}$ around the origin, we have a square area $(-4,4)\times(-4,4)$, and 
three concentric circles with their radiuses being $1$, $2$, and $3$.
Four regions/classes are interlocked and nonconvex, see \cite{sing:nips89} for further 
details of the benchmark. 
Samples are drawn in the box for training with their corresponding outputs being 
the classes $\{1,2,3,4\}$.

We firstly demonstrate theoretical convergence properties of the AN/GGN algorithm
with $\alpha = 1$, by deploying an MLP architecture $\bm{\mathcal{F}}(2,20,1)$.
Dummy units are introduced in both the input and hidden layer.
Activation functions in the hidden layer are chosen to be the \emph{Bend identity}, 
and the error function is the squared loss.
With a set of $20$ randomly drawn training samples, we run the AN/GGN 
algorithm from randomly initialised weights.
The convergence is measured by the distance of the accumulation point 
$\mathbf{W}^{*}$ to all iterates $\mathbf{W}^{(k)}$, i.e., by an extension of
the Frobenius norm of matrices to collections of matrices as 
$\|\mathbf{W}^{(k)}-\mathbf{W}^{*}\|_{F} := \sum_{l=1}^{L}
\|W_{l}^{(k)}-W_{l}^{*}\|_{F}$.
It is clear from Figure~\ref{fig:quad} that the AN/GGN algorithm converges locally 
quadratically fast to a global minimiser of exact learning.

We further investigate the performance of AN/GGN with five different activations, 
namely \emph{Sigmoid}, \emph{$\operatorname{tanh}$}, \emph{SoftSign}, 
\emph{SoftPlus}, and \emph{Bent identity}.
The first three activation functions are squashing, while the SoftPlus is only 
bounded from below, and the Bent identity is totally unbounded.
Figure~\ref{fig:vanish} gives the box plot of the value of the total loss function 
$\mathcal{J}$ over $100$ independent runs from random initialisations after convergence.
Clearly, AN/GGNs with SoftPlus and 
Bent identity perform very well, while the ones with Sigmoid and 
$\operatorname{tanh}$ suffer from numerically spurious local minima.
However, it is also very interesting to observe that SoftSign does not share the bad
convergence behaviour as its squashing counterparts, due to some unknown factors.


Finally, we investigate the AN/GGN algorithm, comparing with the classic BP algorithm 
in terms of convergence speed, without exact learning being assumed. 
In this experiment, we adopt an MLP architecture $\bm{\mathcal{F}}(2,10,10,4)$,
where the target outputs being the corresponding standard basis vector in 
$\mathbb{R}^{4}$, and the set step size to be constant $\alpha = 0.01$.
For running $1000$ iterations, the BP algorithm took $61.1$ sec.,
while the AN/GGN algorithm spent $1314.1$ sec.
On average, the running time for each iteration of AN/GGN was about $21.4$ times as required 
for an iteration of BP.
With the same data and 
the same random initialisation, we ran BP for $20760$ iterations, which took the 
same amount of time as required for $1000$ iterations of AN/GGN.
As shown in Figure~\ref{fig:conv}, the first $1000$ iterations of BP was 
highlighted in \emph{red} with the remaining iterations being coloured in \emph{blue}.
The AN/GGN goes up at the beginning, then smoothly converges to the global minimal 
value, while the BP demonstrates strong oscillation towards the end.
%

%

\section{Conclusion}
\label{sec:07}
In this work, we provide a smooth optimisation perspective on the challenge of 
training MLPs.
Under the condition of exact learning, we characterise the criti\-cal point 
conditions of the empirical total loss function, and investigate sufficient conditions 
to ensure any local minimum to be globally minimal.
Classic results on MLPs with only one hidden layer are reexamined in the proposed framework.
Finally, the so-called Generalised Gauss-Newton algorithm is rigorously revisited as 
an approximate Newton's algorithm, which shares the property of being locally 
quadratically convergent to a global minimum.
All aspects discussed in this paper require a further systematic and thorough investigation
both theoretically and experimentally, and are expected to be also applicable for 
training recurrent neural networks.

\appendix

\section{Tracy-Singh product and Khatri-Rao pro\-duct}
%
Given two matrices $\mathbf{A} \in \mathbb{R}^{m \times n}$ and 
$\mathbf{B} \in \mathbb{R}^{p \times q}$.
Let us partition $\mathbf{A}$ into blocks $A_{ij} \in \mathbb{R}^{m_{i} \times n_{j}}$, 
and $\mathbf{B}$ into blocks $B_{kl} \in \mathbb{R}^{p_{k} \times q_{l}}$.
The \emph{Tracy-Singh product} of $\mathbf{A}$ and $\mathbf{B}$ \cite{trac:sn72}
is defined as
\begin{equation}
\label{eq:tracy-singh}
	\mathbf{A} \circledast \mathbf{B} 
	= (A_{ij} \circledast \mathbf{B})_{ij} 
	= ((A_{ij} \otimes B_{kl})_{kl})_{ij},
\end{equation}
where the notion $(\cdot)_{ij}$ follows the convention of referring to the 
$(i,j)$-th block of a partitioned matrix.
The matrix $\mathbf{A} \circledast \mathbf{B}$ is of the dimension 
$(m p) \times (n q)$, and its rank shares the same property as the 
\emph{Kronecker product} of matrices as
\begin{equation}
	\operatorname{rank}(\mathbf{A} \circledast \mathbf{B}) = 
	\operatorname{rank}(\mathbf{A}) \operatorname{rank}(\mathbf{B}).
\end{equation}
If $\mathbf{A}$ and $\mathbf{B}$ are partitioned identically, then the 
\emph{Khatri-Rao product} of the two matrices is defined as 
\begin{equation}
\label{eq:khatri-rao}
	\mathbf{A} \odot \mathbf{B} 
	= (A_{ij} \otimes B_{ij})_{ij} .
\end{equation}
The matrix $\mathbf{A} \odot \mathbf{B}$ is of the dimension 
$(\sum_{i}\!m_{i} p_{i}) \times (\sum_{i}\!n_{i} q_{i})$.
The connection between the \emph{Tracy-Singh product} and the \emph{Khatri-Rao product}
is given as
\begin{equation}
	\mathbf{A} \odot \mathbf{B} 
	= Z_{1}^{\top} ( \mathbf{A} \circledast \mathbf{B} ) Z_{2},
\end{equation}
where $Z_{1} \!\in\! \mathbb{R}^{(m p) \times (\sum_{i}\!m_{i} p_{i})}$ and 
$Z_{2} \!\in\! \mathbb{R}^{(n q) \times (\sum_{i}\!n_{i} q_{i})}$ are
two selection matrices, satisfying $Z_{1}^{\top} Z_{1} = I_{\sum_{i}\!m_{i} p_{i}}$
and $Z_{2}^{\top} Z_{2} = I_{\sum_{i}\!n_{i} q_{i}}$.
We refer to \cite{lius:laa99} for concrete constructions of matrices $Z_{1}$ and $Z_{2}$,
and more technical details regarding the \emph{Khatri-Rao product}.
It is then trivial to conclude the following corollary.
\begin{corollary}
\label{cor:01}
	Given two identically partitioned matrices $\mathbf{A}$ and $\mathbf{B}$,
	the rank of the \emph{Tracy-Singh product} and the rank of the \emph{Khatri-Rao product} 
	of both matrices fulfils the following inequality
    \begin{equation}
    	\operatorname{rank}(\mathbf{A} \odot \mathbf{B})
		\le
		\operatorname{rank}(\mathbf{A} \circledast \mathbf{B}).
    \end{equation}
\end{corollary} 
Furthermore, it is clear that 
\begin{equation}
	\operatorname{rank}(Z_{1}) = \sum_{i}\!m_{i} p_{i},
\end{equation} 
and
\begin{equation}
	\operatorname{rank}(Z_{2}) = \sum_{i}\!n_{i} q_{i}.
\end{equation}

Now, we recall the \emph{Frobenius' rank inequality} \cite{simo:book12},
i.e., given three matrices $A$, $B$, $C$ that have compatible dimensions, then
\begin{equation}
	\operatorname{rank}(A B C) + \operatorname{rank}(B)
	\ge
	\operatorname{rank}(A B) + \operatorname{rank}(B C). \!
\end{equation}
%
%
A special case of the \emph{Frobenius' rank inequality} is the so-called
\emph{Sylvester's rank inequality}, i.e., given two matrices 
$A \in \mathbb{R}^{m \times n}$ and $C \in \mathbb{R}^{n \times p}$,
then the rank of the product of $U$ and $V$ is bounded by 
\begin{equation}
	\operatorname{rank}(A C) \ge \operatorname{rank}(A) + \operatorname{rank}(C) - n.
\end{equation}
%
%
Let $A \in \mathbb{R}^{m \times n_{1}}$,
$B \in \mathbb{R}^{n_{1} \times n_{2}}$, and $C \in \mathbb{R}^{n_{2} \times p}$.
By combining both the Frobenius' rank inequality and the Sylvester's rank inequality,
we have
\begin{equation}
\label{eq:abc}
\begin{split}
	\operatorname{rank}(A B C) 
	\ge &~\!
	\operatorname{rank}(A B) + \operatorname{rank}(B C) - \operatorname{rank}(B)\! \\
	\ge &~\! \operatorname{rank}(A) + \operatorname{rank}(B) - n_{1} +
	 \operatorname{rank}(B) + \operatorname{rank}(C) - n_{2} ~- \\
	 & - \operatorname{rank}(B) \\
	= &~\! \operatorname{rank}(A) + \operatorname{rank}(B) + \operatorname{rank}(C) 
		- n_{1} - n_{2}.
\end{split}
\end{equation}

If the \emph{Tracy-Singh product} $\mathbf{A} \circledast \mathbf{B}$ has full rank, denoted 
by
\begin{equation}
	R_{ts} := \operatorname{rank}(\mathbf{A} \circledast \mathbf{B}),
\end{equation}
then the rank of the \emph{Khatri-Rao product} $\mathbf{A} \odot \mathbf{B}$
is bounded from below by
\begin{equation}
\begin{split}
	\operatorname{rank}(\mathbf{A} \odot \mathbf{B}) \ge &
	\sum_{i}\!m_{i} p_{i} + R_{ts} + \sum_{j}\!n_{j} q_{j} - mp - nq.
\end{split}
\end{equation}
Note, that the above lower bound is not guaranteed to be positive.
Hence, nothing is conclusive about the rank of the \emph{Khatri-Rao product} of 
two arbitrary full rank matrices.

\section{Proof of Proposition~\ref{prop:row_rank}}
%
\begin{proof}
	We can trivially rewrite the Kronecker product for each partition as
	\begin{equation}
	\begin{split}
		\Psi_{i} \otimes \phi_{i} = &~(I_{n_{l}}\Psi_{i}) \otimes (\phi_{i} 1) \\[0.5mm]
		= &~(I_{n_{l}} \otimes \phi_{i}) \Psi_{i}.
	\end{split}
	\end{equation}
	Then, the \emph{Khatri-Rao product} of $\mathbf{\Psi}$ and $\mathbf{\Phi}$ can be computed 
	as the product of two matrices, i.e.,
	\begin{equation}
		\mathbf{\Psi} \odot \mathbf{\Phi} \!= \!\!\!\!\underbrace{[I_{n_{l}} \!\otimes\! 
		\phi_{1}, \ldots, 
		I_{n_{l}} \!\otimes\! \phi_{T}	]}_{=: (I_{n_{l}} \circledast \mathbf{\Phi}) \in
		\mathbb{R}^{(n_{l} n_{n-1}) \times (n_{l} T)}} \!
		\underbrace{\operatorname{diag}(\Psi_{1}, \ldots, \Psi_{T})
		}_{=:\widetilde{\mathbf{\Psi}} \in \mathbb{R}^{(n_{l} T) \times (n_{L} T)}}, \!
	\end{equation}
	where $I_{n_{l}} \circledast \mathbf{\Phi}$ denotes the \emph{Tracy-Singh product} of
	the identity matrix $I_{n_{l}}$ and $T$ column-wised partitioned matrix $\mathbf{\Phi}$,
	and the operator $\operatorname{diag}(\cdot)$ puts a sequence of matrices into 
	a block diagonal matrix.
	By the rank property of the \emph{Tracy-Singh product}, the rank of matrix 
	$I_{n_{l}} \circledast \mathbf{\Phi}$ is equal to $n_{l} 
	\operatorname{rank}(\mathbf{\Phi})$.
	Further, by the \emph{Sylvester's rank inequality}, the rank of 
	$\mathbf{\Psi} \odot \mathbf{\Phi}$ is bounded from below 
	\begin{equation}
		\operatorname{rank}(\mathbf{\Psi} \odot \mathbf{\Phi})
		\ge
		n_{l} \operatorname{rank}(\mathbf{\Phi}) + 
		\sum\limits_{i=1}^{T}\operatorname{rank}(\Psi_{i}) - T n_{L}.
	\end{equation}
	Specifically, if all matrices $\Psi_{i}$'s and $\mathbf{\Phi}$ are of full rank,
	we have the following properties.
	\begin{enumerate}[(1)]
		\item If $n_{l} \le n_{L}$, then the rank of the block diagonal matrix 
			$\widetilde{\mathbf{\Psi}}$ is equal to $n_{l} T$. 
			By the Sylvester's rank inequality \cite{simo:book12}, we have
			\begin{equation}
			\begin{split}
				\operatorname{rank}(\mathbf{\Psi} \odot \mathbf{\Phi}) \ge &~ n_{l}  
				\operatorname{rank}(\mathbf{\Phi}) + n_{l} T - n_{l} T \\
				= &~ n_{l} \operatorname{rank}(\mathbf{\Phi}).
			\end{split}
			\end{equation}
			
		\item If $n_{l} > n_{L}$ and $n_{l-1} \ge T$, then the rank of 
			$\widetilde{\mathbf{\Psi}}$ is equal to	$n_{L} T$, and the rank of 
			$(I_{n_{l}} \circledast \mathbf{\Phi})$
			is equal to $n_{l} T$.
			By the Sylvester's rank inequality, we have
			\begin{equation}
			\begin{split}
				\operatorname{rank}(\mathbf{\Psi} \odot \mathbf{\Phi}) \ge &~ 
				n_{l} T + n_{L} T - n_{l} T \\
				= &~ n_{L} T.
			\end{split}
			\end{equation}
		\item If $n_{l} > n_{L}$ and $n_{l-1} < T$, then the rank of 
			$(I_{n_{l}} \circledast \mathbf{\Phi})$
			is equal to $n_{l} n_{l-1}$. By the same argument, we have
			\begin{equation}
				\operatorname{rank}(\mathbf{\Psi} \odot \mathbf{\Phi}) \ge 
				n_{l} n_{l-1} + n_{L} T - n_{l} T.
			\end{equation}
			It is clear that such a lower bound can be even nega\-tive, i.e.,
			practically useless.
			However, since matrix $\mathbf{\Phi}$ is of full rank, there 
			must exist a non-zero vector $\phi_{i}$, so that 
			$\operatorname{rank}(\Psi_{i} \otimes \phi_{i}) = n_{L}$.
			Then we have the result $\operatorname{rank}(\mathbf{\Psi} \odot 
			\mathbf{\Phi}) \ge n_{L}$.
	\end{enumerate}
	\vspace{-6.3mm}
\end{proof}

\section{Proof of Proposition~\ref{prop:rank_jacobi}}
\begin{proof}
	By stacking all row blocks $\mathbf{\Psi}_{l} \odot \mathbf{\Phi}_{l-1}$ 
	for $l=1, \ldots, L$ together, we have $\mathbf{P}(\mathbf{W})$ as 
	in Eq.~\eqref{eq:jacobi}.
	We can rewrite $\mathbf{P}(\mathbf{W})$ as
	\begin{equation}
	\begin{split}
		\mathbf{P}(\mathbf{W}) = & \operatorname{diag}\big(
		I_{n_{1}} \!\!\circledast\! \mathbf{\Phi}_{0}, \ldots,
		I_{n_{L}} \!\!\circledast\! \mathbf{\Phi}_{L-1}\big) 
		\cdot \operatorname{diag}\big(
		\widetilde{\mathbf{\Psi}}_{1}, \ldots, \widetilde{\mathbf{\Psi}}_{L}\big)
		\cdot \mathbf{I}_{T n_{l}}^{L},
	\end{split}
	\end{equation}
	where $\mathbf{I}_{T n_{l}}^{L} := [I_{T n_{L}}, \ldots, I_{T n_{L}}]^{\top}
	\in \mathbb{R}^{LTn_{L} \times Tn_{L}}$ is a matrix of stacking $L$ copies of 
	the identity matrix $I_{T n_{L}}$ on top of each other.
	Then, by applying Eq.~\eqref{eq:abc}, it is straightforward to get
	\begin{equation}
	\begin{split}
		\operatorname{rank}\!\big(\mathbf{P}(\mathbf{W})\big) 
		\ge & \sum\limits_{l=1}^{L} n_{l} \operatorname{rank}\big(\mathbf{\Phi}_{l-1}\big) 
		+ \sum\limits_{l=1}^{L} \sum\limits_{i=1}^{T} 
		\operatorname{rank}\big(\Psi_{l}^{(i)}\big) + T n_{L} ~- \\
		& - \sum\limits_{l=1}^{L} T n_{l} - L T n_{L}.
	\end{split}
	\end{equation}
	The result follows directly.
\end{proof}

It is clear that such a bound in Proposition~\ref{prop:rank_jacobi} is still very 
problem-dependent, and hard to control.
Nevertheless, due to the special structure of $\mathbf{I}_{T n_{l}}^{L}$,
the actual rank bound is given practically by the largest bound of each individual 
row block as characterised in Proposition~\ref{prop:row_rank}, i.e.,
\begin{equation}
		\operatorname{rank}\!\big(\mathbf{P}(\mathbf{W})\big) 
		\ge \max_{1 \le l \le L} \operatorname{rank}\big( 
		\mathbf{\Psi}_{l} \odot \mathbf{\Phi}_{l-1} \big).
\end{equation}

\section{Proof of Proposition~\ref{prop:hidden_t}}
\begin{proof}
	We feed samples $X := [x_{1}, \ldots, x_{T}] \in \mathbb{R}^{n_{0} \times T}$ 
	through the MLP to generate the outputs in the hidden layer 
	$\Phi_{1} := [\phi_{1}^{(1)}, \ldots, \phi_{1}^{T}] \in \mathbb{R}^{T \times T}$, 
	which is invertible due to Condition (2). 
	It can be achieved by employing appropriate activation functions as
	suggested in \cite{itoy:acm96}, such as the \emph{Sigmoid} and the 
	$\operatorname{tanh}$.
	Then in the output layer, we have 
	$\Phi_{2} := [\phi_{2}^{(1)}, \ldots, \phi_{2}^{(T)}] = 
	W_{2}^{\top} \Phi_{1} \in \mathbb{R}^{n_{2} \times T}$.
	Let us denote by $Y := [g^{*}(x_{1}), \ldots, g^{*}(x_{T})] \in 
	\mathbb{R}^{n_{2} \times T}$ the desired outputs.
	Then, every pair $(W_{1}, (Y \Phi_{1}^{-1})^{\top})$ is a global minimum of the 
	total loss function.

	We then compute the critical point conditions in the output layer as 
	\begin{equation}
		\underbrace{\!
    	\left[\!\!\! \begin{array}{ccc}
    	I_{n_{2}} \!\otimes \phi_{1}^{(1)} &
    	\!\!\!\!\!\ldots\!\!\!\!\! &
    	I_{n_{2}} \!\otimes \phi_{1}^{(T)}
    	\end{array}\!\!\!\right]\!
		}_{:= P_{2} \in \mathbb{R}^{(T n_{2}) \times (T n_{2})}}
		\left[\!\!\! \begin{array}{c}
    	\nabla_{\!E}(\phi_{2}^{(1)}) \\
    	\vdots \\
    	\nabla_{\!E}(\phi_{2}^{(T)})
    	\end{array}\!\!\!\right] \!= 0,
    \end{equation}
	where $P_{2}$ is a square matrix.
	By following case (1) in Proposition~\ref{prop:row_rank}, we get
	%
		$\operatorname{rank}(P_{2}) = T n_{2}$.
	%
	The result simply follows.
\end{proof}

Note, that in Proposition~\ref{prop:hidden_t}, we do not consider the dummy units 
introduced by the scalar-valued bias $b_{l,k}$.
Nevertheless, using similar arguments, the statements in Proposition~\ref{prop:hidden_t}
also hold true for the case with free variables $b_{l,k}$.

\section{Proof of Proposition~\ref{prop:hidden_s}}
\begin{proof}
	It is straightforward to have
	\begin{equation}
	\begin{split}
		\operatorname{rank}\!\big(\mathbf{P}(\mathbf{W})\big) 
		\ge &~n_{1} \operatorname{rank}\big(\mathbf{\Phi}_{0}\big) +
		n_{2} n_{1} - T n_{1} + 2 T n_{2} - 2 T n_{2} \\
		= &~ n_{1} \big(\operatorname{rank}\big(\mathbf{\Phi}_{0}\big) +
		n_{2} - T \big).
	\end{split}
	\end{equation}
	Proposition~\ref{prop:row_rank} implies
	\begin{equation}
		\operatorname{rank}(\mathbf{\Psi}_{2} \odot \mathbf{\Phi}_{1})
		\ge n_{2} \operatorname{rank}(\mathbf{\Phi}_{1}),
	\end{equation}
	and
	\begin{equation}
		\operatorname{rank}(\mathbf{\Psi}_{1} \odot \mathbf{\Phi}_{0})
		\ge n_{2}.
	\end{equation}
	By the construction of $\operatorname{rank}(\mathbf{\Phi}_{1}) \ge n_{1}$, 
	the result follows directly.
\end{proof}


\begin{thebibliography}{10}
\providecommand{\url}[1]{#1}
\csname url@samestyle\endcsname
\providecommand{\newblock}{\relax}
\providecommand{\bibinfo}[2]{#2}
\providecommand{\BIBentrySTDinterwordspacing}{\spaceskip=0pt\relax}
\providecommand{\BIBentryALTinterwordstretchfactor}{4}
\providecommand{\BIBentryALTinterwordspacing}{\spaceskip=\fontdimen2\font plus
\BIBentryALTinterwordstretchfactor\fontdimen3\font minus
  \fontdimen4\font\relax}
\providecommand{\BIBforeignlanguage}[2]{{%
\expandafter\ifx\csname l@#1\endcsname\relax
\typeout{** WARNING: IEEEtran.bst: No hyphenation pattern has been}%
\typeout{** loaded for the language `#1'. Using the pattern for}%
\typeout{** the default language instead.}%
\else
\language=\csname l@#1\endcsname
\fi
#2}}
\providecommand{\BIBdecl}{\relax}
\BIBdecl

\bibitem{bish:book96}
C.~M. Bishop, \emph{Neural Networks for Pattern Recognition}.\hskip 1em plus
  0.5em minus 0.4em\relax Oxford University Press, USA, 1996.

\bibitem{lecu:nature15}
Y.~LeCun, Y.~Bengio, and G.~Hinton, ``Deep learning,'' \emph{Nature}, vol. 521,
  pp. 436--444, 2015.

\bibitem{yudo:book15}
D.~Yu and L.~Deng, \emph{Automatic Speech Recognition: A Deep Learning
  Approach}.\hskip 1em plus 0.5em minus 0.4em\relax Springer-Verlag, London,
  2015.

\bibitem{glor:aistats10}
X.~Glorot and Y.~Bengio, ``Understanding the difficulty of training deep
  feedforward neural networks,'' in \emph{Proceedings of the $13^{th}$
  International Conference on Artificial Intelligence and Statistics
  (AISTATS)}, vol.~9, 2010, pp. 249--256.

\bibitem{horn:nn91}
K.~Hornik, ``Approximation capabilities of multilayer feedforward networks,''
  \emph{Neural Networks}, vol.~4, no.~2, pp. 251--257, 1991.

\bibitem{suns:aaai16}
S.~Sun, W.~Chen, L.~Wang, X.~Liu, and T.-Y. Liu, ``On the depth of deep neural
  networks: A theoretical view,'' in \emph{Proceedings of the $31^{st}$ AAAI
  Conference on Artificial Intelligence}, 2016, pp. 2066--2072.

\bibitem{mhas:nips93}
H.~N. Mhaskar and C.~A. Micchelli, ``How to choose an activation function,'' in
  \emph{Proceedings of the 6th International Conference on Neural Information
  Processing Systems}, 1993, pp. 319--326.

\bibitem{fala:ijcnn99}
T.~Falas and A.~G. Stafylopatis, ``The impact of the error function selection
  in neural network-based classifiers,'' in \emph{Proceedings of the
  International Joint Conference on Neural Networks (IJCNN)}, vol.~3, 1999, pp.
  1799--1804.

\bibitem{widr:pieee90}
B.~Widrow and M.~A. Lehr, ``30 years of adaptive neural networks: perceptron,
  madaline, and backpropagation,'' \emph{Proceedings of the IEEE}, vol.~78,
  no.~9, pp. 1415--1442, 1990.

\bibitem{sutt:ccss86}
R.~S. Sutton, ``Two problems with backpropagation and other steepest-descent
  learning procedures for networks,'' in \emph{Proceedings of the $8$-th Annual
  Conference of the Cognitive Science Society}, 1986, pp. 823--831.

\bibitem{roja:book96}
R.~Rojas, \emph{Neural Networks: A Systematic Introduction}.\hskip 1em plus
  0.5em minus 0.4em\relax Springer, 1996.

\bibitem{hame:nn98}
L.~G.~C. Hamey, ``{XOR} has no local minima: A case study in neural network
  error surface analysis,'' \emph{Neural Metworks}, vol.~11, no.~4, pp.
  669--681, 1998.

\bibitem{spri:amai99}
I.~G. Sprinkhuizen-Kuyper and E.~J.~W. Boers, ``The local minima of the error
  surface of the 2-2-1 {XOR} network,'' \emph{Annals of Mathematics and
  Artificial Intelligence}, vol.~25, no.~1, pp. 107--136, 1999.

\bibitem{chor:aistats15}
A.~Choromanska, M.~Henaff, M.~Mathieu, G.~B. Arous, and Y.~Le{C}un, ``The loss
  surfaces of multilayer networks,'' in \emph{Proceedings of the $18^{th}$
  International Conference on Artificial Intelligence and Statistics
  (AISTATS)}, 2015, pp. 192--204.

\bibitem{lisb:network91}
P.~J.~G. Lisboa and S.~J. Perantonis, ``Complete solution of the local minima
  in the {XOR} problem,'' \emph{Network: Computation in Neural Systems},
  vol.~2, no.~1, pp. 119--124, 1991.

\bibitem{spri:tnn99}
I.~G. Sprinkhuizen-Kuyper and E.~J.~W. Boers, ``A local minimum for the 2-3-1
  {XOR} network,'' \emph{IEEE Transactions on Neural Networks}, vol.~10, no.~4,
  pp. 968--971, 1999.

\bibitem{gori:pami92}
M.~Gori and A.~Tesi, ``On the problem of local minima in backpropagation,''
  \emph{IEEE Transactions on Pattern Analysis and Machine Intelligence},
  vol.~14, no.~1, pp. 76--86, 1992.

\bibitem{yuxi:tnn92}
X.-H. Yu, ``Can backpropagation error surface not have local minima,''
  \emph{IEEE Transactions on Neural Networks}, vol.~3, no.~6, pp. 1019--1021,
  1992.

\bibitem{yuxi:tnn95}
X.-H. Yu and G.-A. Chen, ``On the local minima free condition of
  backpropagation learning,'' \emph{IEEE Transactions on Neural Networks},
  vol.~6, no.~5, pp. 1300--1303, 1995.

\bibitem{kole:cs90}
J.~F. Kolen and J.~B. Pollack, ``Backpropagation is sensitive to initial
  conditions,'' \emph{Complex Systems}, vol.~4, no.~3, pp. 269--280, 1990.

\bibitem{good:iclr15}
I.~J. Goodfellow, O.~Vinyals, and A.~M. Saxe, ``Qualitatively characterizing
  neural network optimization problems,'' Published at the $5^{th}$
  International Conference on Learning Representations (ICLR).
  arXiv:1412.6544., 2015.

\bibitem{kawa:nips16}
K.~Kawaguchi, ``Deep learning without poor local minima,'' in \emph{Advances in
  Neural Information Processing Systems 29}, D.~D. Lee, M.~Sugiyama, U.~V.
  Luxburg, I.~Guyon, and R.~Garnett, Eds., 2016, pp. 586--594.

\bibitem{nguy:icml17}
Q.~Nguyen and M.~Hein, ``The loss surface of deep and wide neural networks,''
  in \emph{Proceedings of the $34^{th}$ International Conference on Machine
  Learning}, 2017.

\bibitem{haef:cvpr17}
B.~D. Haeffele and R.~Vidal, ``Global optimality in neural network training,''
  in \emph{Proceedings of the IEEE Conference on Computer Vision and Pattern
  Recognition (CVPR)}, 2017, pp. 7331 -- 7339.

\bibitem{vogl:bc88}
T.~P. Vogl, J.~K. Mangis, A.~K. Rigler, W.~T. Zink, and D.~L. Alkon,
  ``Accelerating the convergence of the back-propagation method,''
  \emph{Biological Cybernetics}, vol.~59, no.~4, pp. 257--263, 1988.

\bibitem{char:ieepg92}
C.~Charalambous, ``Conjugate gradient algorithm for efficient training of
  artificial neural networks,'' \emph{IEE Proceedings G - Circuits, Devices and
  Systems}, vol. 139, no.~3, pp. 301--310, 1992.

\bibitem{lequ:icml11}
Q.~Le, J.~Ngiam, A.~Coates, A.~Lahiri, B.~Prochnow, and A.~Ng, ``On
  optimization methods for deep learning,'' Proceedings of international
  conference on Machine Learning, 2011.

\bibitem{king:iclr15}
D.~P. Kingma and J.~L. Ba, ``Adam: a method for stochastic optimization,'' in
  \emph{The $3$-rd International Conference on Learning Representations}, 2015,
  pp. 1--13.

\bibitem{schr:nc02}
N.~N. Schraudolph, ``Fast curvature matrix-vector products for second-order
  gradient descent,'' \emph{Neural Computation}, vol.~14, pp. 1723 -- 1738,
  2002.

\bibitem{mart:icml10}
J.~Martens, ``Deep learning via {H}essian-free optimization,'' in
  \emph{Proceedings of the $27^{th}$ International Conference on Machine
  Learning (ICML)}, 2010, pp. 735--742.

\bibitem{viny:aistats12}
O.~Vinyals and D.~Povey, ``Krylov subspace descent for deep learning,'' in
  \emph{Proceedings of the $15^{th}$ International Conference on Arti cial
  Intelligence and Statistics (AISTATS)}, vol.~22, 2012, pp. 1261--1268.

\bibitem{mart:arxiv14}
J.~Martens, ``New perspectives on the natural gradient method,''
  arXiv:1412.1193v8, 2014.

\bibitem{batt:nc92}
R.~Battiti, ``First- and second-order methods for learning: Between steepest
  descent and newton's method,'' \emph{Neural Computation}, vol.~4, no.~2, pp.
  141--166, 1992.

\bibitem{wang:nn98}
Y.-J. Wang and C.-T. Lin, ``A second-order learning algorithm for multilayer
  networks based on block {H}essian matrix,'' \emph{Neural Networks}, vol.~11,
  no.~9, pp. 1607--1622, 1998.

\bibitem{bish:nc92}
C.~Bishop, ``Exact calculation of the hessian matrix for the multilayer
  perceptron,'' \emph{Neural Computation}, vol.~4, no.~4, pp. 494--501, 1992.

\bibitem{mizu:nn08}
E.~Mizutani and S.~E. Dreyfus, ``Second-order stagewise backpropagation for
  hessian-matrix analyses and investigation of negative curvature,''
  \emph{Neural Networks}, vol.~21, no. 2-3, pp. 193--203, 2008.

\bibitem{guel:book10}
O.~G{\"u}ler, \emph{Foundations of Optimization}.\hskip 1em plus 0.5em minus
  0.4em\relax Springer, 2010.

\bibitem{itoy:acm96}
Y.~Ito, ``Nonlinearity creates linear independence,'' \emph{Advances in
  Computational Mathematics}, vol.~5, no.~1, pp. 189--203, 1996.

\bibitem{shah:tnn99}
J.~V. Shah and C.-S. Poon, ``Linear independence of internal representations in
  multilayer perceptrons,'' \emph{IEEE Transactions on Neural Networks},
  vol.~10, no.~1, pp. 10--18, 1999.

\bibitem{hart:book04}
H.~R. and A.~Zisserman, \emph{Multiple View Geometry in Computer Vision}.\hskip
  1em plus 0.5em minus 0.4em\relax Cambridge University Press, New York, 2004.

\bibitem{simo:book12}
D.~A. Simovici, \emph{Linear Algebra Tools for Data Mining}.\hskip 1em plus
  0.5em minus 0.4em\relax World Scientific Publishing Company, 2012.

\bibitem{miln:book63}
J.~Milnor, \emph{Morse Theory}.\hskip 1em plus 0.5em minus 0.4em\relax
  Princeton University Press, 1963.

\bibitem{fair:nc12}
M.~Fairbank and E.~Alonso, ``Efficient calculation of the {G}auss-{N}ewton
  approximation of the {H}essian matrix in neural networks,'' \emph{Neural
  Computation}, vol.~24, no.~3, pp. 607--610, 2012.

\bibitem{bote:icml17}
A.~Botev, H.~Ritter, and D.~Barber, ``Practical gauss-newton optimisation for
  deep learning,'' in \emph{Proceedings of the $34^{th}$ International
  Conference on Machine Learning}, 2017.

\bibitem{sing:nips89}
S.~Singhal and L.~Wu, ``Training multilayer perceptrons with the extended
  {K}alman algorithm,'' in \emph{Advances in Neural Information Processing
  Systems}, 1989, pp. 133--140.

\bibitem{trac:sn72}
D.~S. Tracy and R.~P. Singh, ``A new matrix product and its applications in
  partitioned matrices,'' \emph{Statistica Neerlandica}, vol.~26, pp. 143--157,
  1972.

\bibitem{lius:laa99}
S.~Liu, ``Matrix results on the {K}hatri-{R}ao and {T}racy-{S}ingh products,''
  \emph{Linear Algebra and its Applications}, vol. 289, no.~1, pp. 267--277,
  1999.

\end{thebibliography}


\end{document}